\pdfoutput=1

\documentclass[11pt]{article}

\usepackage[final]{acl}

\usepackage{times}
\usepackage{latexsym}

\usepackage[T1]{fontenc}

\usepackage[utf8]{inputenc}
\usepackage{longtable}
\usepackage{microtype}
\usepackage[utf8]{inputenc}
\usepackage{inconsolata}

\usepackage{graphicx}
\usepackage{amsmath}
\usepackage{tikz}
\usepackage{hyperref}

\usepackage{geometry}
\usepackage{booktabs} 
\usepackage{booktabs} 
\usepackage{caption}
\usepackage{multirow} 
\usepackage{xcolor}   
\usepackage{amssymb}  
\usepackage{pifont}   
\usepackage{float}
\usepackage{array}     
\usepackage{url}
\usepackage{tikz}
\usepackage{adjustbox} 
\usepackage{graphicx}
\usepackage{subcaption} 
\usepackage{algorithm}
\usepackage{mathrsfs}
\usepackage{algorithmic}
\usepackage{amsmath}
\usepackage{amssymb}
\usepackage{mathrsfs} 
\usepackage{enumitem}

\title{Instruct-of-Reflection: Enhancing Large Language Models Iterative Reflection Capabilities via Dynamic-Meta Instruction}

\author{
  \textbf{Liping Liu\textsuperscript{1}\thanks{Work done while Liping Liu was an intern at AI Lab of Lenovo Research. }}, 
  \textbf{Chunhong Zhang\textsuperscript{1}}, 
  \textbf{Likang Wu\textsuperscript{3}}, 
  \textbf{Chuang Zhao}\textsuperscript{4},\\ 
  \textbf{Zheng Hu\textsuperscript{1}\thanks{Corresponding authors.}}, 
  \textbf{Ming He\textsuperscript{2†}}, 
  \textbf{Jianping Fan\textsuperscript{2}} \\
  \textsuperscript{1}\thanks{The first affiliation is State Key Laboratory of Networking and Switching Technology, Beijing University of Posts and Telecommunications, Beijing 100876, China.}Beijing University of Posts and Telecommunications \\
  \textsuperscript{2}AI Lab of Lenovo Research   
  \textsuperscript{3}Tianjin University \\
  \textsuperscript{4}The Hong Kong University of Science and Technology \\
  \texttt{{\{liuliping, Huzheng,  zhangch\}}@bupt.edu.cn},
  \texttt{heming01@foxmail.com}\\ \texttt{wulk@tju.edu.cn}, \texttt{czhaobo@connect.ust.hk}, \texttt{jfan1@lenovo.com}
}

\begin{document}
\maketitle

\begin{abstract}
Self-reflection for Large Language
Models (LLMs) has gained significant attention. Existing approaches involve models iterating and improving their previous responses based on LLMs' internal reflection ability or external feedback. However, recent research has raised doubts about whether intrinsic self-correction without external feedback may even degrade performance. Based on our empirical evidence, we find that current static reflection methods may lead to redundant, drift, and stubborn issues.  To mitigate this, we introduce \textbf{I}nstruct-\textbf{o}f-\textbf{R}eflec\textbf{t}ion (\textbf{IoRT}), a novel and general reflection framework that leverages dynamic-meta instruction to enhance the iterative reflection capability of LLMs. Specifically, we propose the instructor driven by the meta-thoughts and self-consistency classifier, generates various instructions, including refresh, stop, and select, to guide the next reflection iteration. Our experiments demonstrate that IoRT achieves an average improvement of 10.1\% over established baselines in mathematical and commonsense reasoning tasks, highlighting its efficacy and applicability. Our code is available at \url{https://github.com/llp635/IoRT}.
\end{abstract}

\section{Introduction}
Large language models (LLMs) \cite{touvron2023llama, achiam2023gpt, anil2023palm, TheC3} have demonstrated remarkable capabilities across various natural language processing tasks \cite{bai2022training, wei2022emergent, chu2023survey}, particularly when leveraging a range of prompting strategies such as Chain-of-Thought \cite{wei2022chain} which improve the reasoning ability of LLMs without increasing model parameters \cite{rae2021scaling, srivastava2023beyond, yin-etal-2023-exchange}. Inspired by how humans iteratively refine their written text \cite{10.5555/3666122.3668141}, the concept of "self-correction" \cite{welleck2023generating} has been proposed and garnered extensive attention where LLMs first generate an initial response, then gather feedback to refine previous responses \cite{ganguli2023capacity, xi2023self, paul2024refiner}. However, recent studies \cite{huanglarge, zhang-etal-2024-self-contrast} raise two doubts about the self-correction capabilities of LLMs: First, \cite{kim2024language, shinn2023reflexion} use oracle labels regarding the answer correctness to guide the self-correction process. However, in practice, the availability of oracle labels seems counter-intuitive because there seems to be little reason to deploy LLMs for problem solving if the ground truth is already known. Second, without any external or human feedback, the performance after LLMs’ intrinsic self-correction even deteriorates. 
\begin{figure}[t]
\centering
\includegraphics[width=1\linewidth]{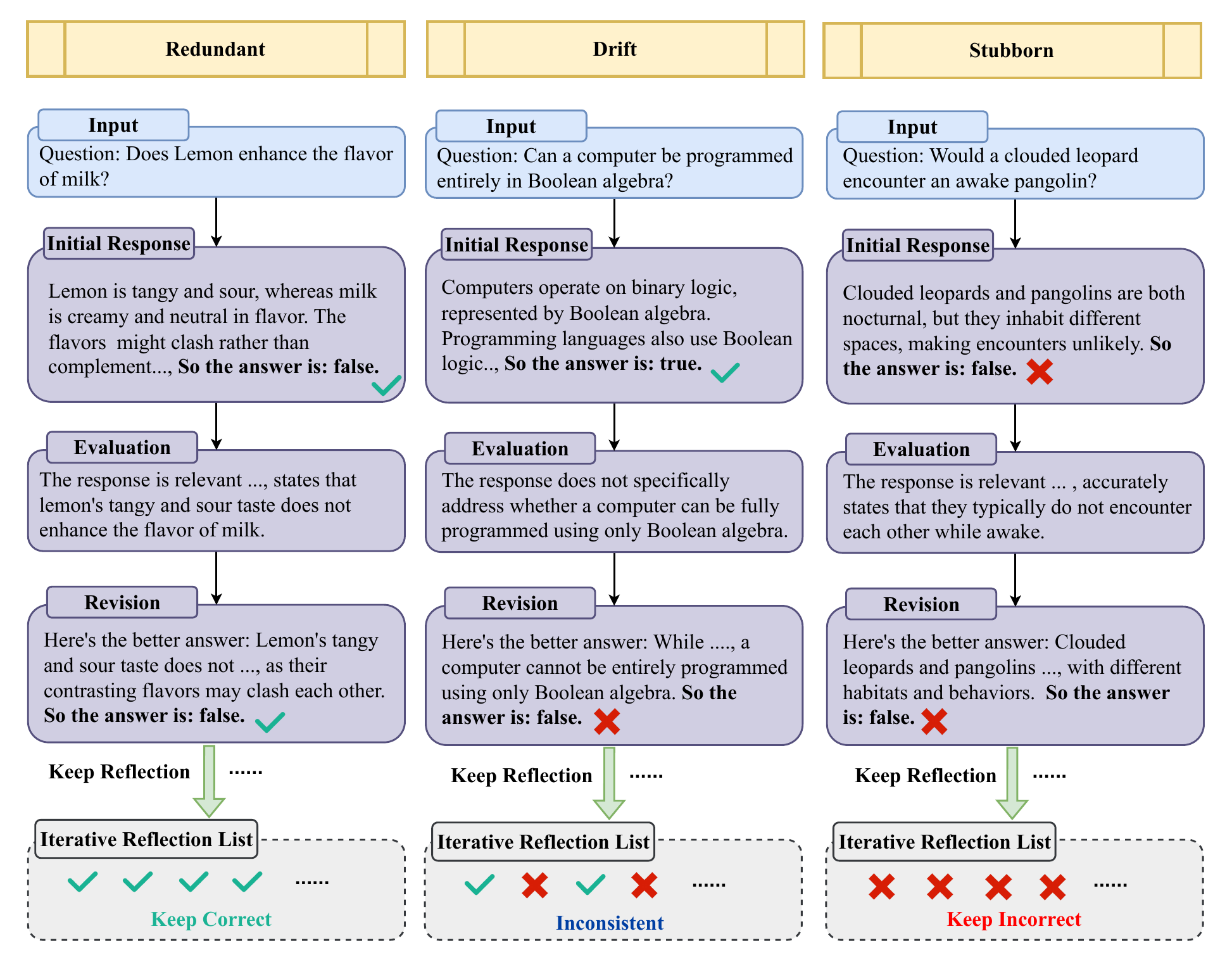}
\caption{Prior reflection research is primarily based on static iterative strategies, which face issues of redundancy, drift, and stubborn.}
\label{fig:1}
\end{figure}

Our research also focuses on the reflective capabilities of LLMs. Unlike previous studies that primarily investigate performance changes in intrinsic reflection, we explore whether the performance of LLMs in reflection deteriorates when external feedback is available. Therefore, in our exploration experiments, we evaluate the performance of CRITIC \cite{gou2024critic}, which leverages the results of tool interactions as external feedback, enabling a systematic investigation of how this feedback impacts the performance of reflection. Our findings (\S~\ref{subsec:Performance of Iterative Reflections}) indicate that neither intrinsic self-correction nor CRITIC performs satisfactorily. To investigate the reasons behind that, we comprehensively analyze the performance across multiple iterations of reflection, which is different from prior research focusing on a single reflection. Figure \ref{fig:1} shows that the $i_{\text{th}}$ reflection is derived from the $(i-1)_{\text{th}}$ response, this static iterative reflection leads to three issues: 
1) \textbf{Redundant} (\textcolor{green}{\ding{51}}$\Rightarrow$\textcolor{green}{\ding{51}}): Although it does not alter accuracy in iterations, it does increase overhead.
2) \textbf{Drift} (\textcolor{green}{\ding{51}}$\Rightarrow$\textbf{\textcolor{red}{\ding{55}}}): Self-correction is highly random, undermining both accuracy and reliability.
3) \textbf{Stubborn} (\textbf{\textcolor{red}{\ding{55}}}$\Rightarrow$\textbf{\textcolor{red}{\ding{55}}}): LLMs consistently persist in errors, failing to properly identify and correct potential mistakes. Additionally, we found that the prevalence of these three issues varies across different LLMs and tasks.

As a remedy to these challenges, we propose Instruct of Reflection (IoRT), an efficient and dynamic iterative reflection framework designed to continuously optimize the reflection process. The framework including three stages: (i) meta-thinker generates meta-thoughts, (ii) refreshing to generate an initial response, and reflector to self-reflect based on feedback. (iii) instruct the iteration. Notably, aiming to mitigate redundant, drift, and stubborn issues, in the third stage, the instructor, augmented by meta-thought and driven by the self-consistency classifier, generates refresh, stop, or select instructions to direct the next reflection iteration. Specifically, when the Instructor determines that both responses have correctly addressed the problem, it generates a stop instruction to promptly halt the reflection process, effectively reducing redundancy. If the instructor assesses that both answers are incorrect, it generates a refresh instruction, helping the LLMs prevent stubbornness. In cases where answers are inconsistent, the instructor selects the more optimal response as the output of the current iteration and the foundation for the next reflection, thereby reducing drift and improving performance accuracy.

We evaluate the performance of IoRT across various LLMs and two distinct tasks: mathematical reasoning and commonsense reasoning. The result demonstrates that IoRT outperforms baselines on established strong baselines, underscoring the critical role of the instructor in augmenting the iterative reflection capabilities of LLMs. Our main contributions can be summarized as follows:

\begin{itemize}[left=0pt]
    \item Our exploration experiments reveal that LLMs struggle to correct previous errors. The static iterative reflection suffers from redundant, drift, and stubborn, which may increase overhead or even deteriorate performance.
    \item We propose Instruct-of-Reflection (IoRT), a dynamic iterative reflection framework that integrates abstract reasoning into the reflection process, adaptively generating different instruction to regulate the reflection process.
    \item Experimental results on three datasets and five different LLMs underscore the efficacy, overhead efficiency, and generality of IoRT. Additionally, without relying on any oracle labels, IoRT demonstrates its practicality.
\end{itemize}

\section{Related Works}
\subsection{Prompt-based Reasoning}  
 To exploit the reasoning ability in LLMs, \cite{wei2022chain} proposes Chain-of-Thought (CoT) prompting which guides the model to generate a series of text describing reasoning steps before reaching the final answer. Subsequently, \cite{chen2022program} introduces Program-of-Thoughts (PoT) prompting which uses llms to generate programming language statements. In recent years, the self-correction capabilities of LLMs have garnered significant attention, with LLMs expected to iteratively improve their previous responses based on feedback \cite{bai2022constitutionalaiharmlessnessai}. Existing studies on self-correction can be categorized into two types based on the feedback source: external feedback and internal feedback \cite{pan2024automatically}. Internal feedback relies on the model’s inherent knowledge and parameters, while external feedback involves inputs from humans, other models \cite{wang2023shepherd, paul-etal-2024-refiner}, or external tools and knowledge sources \cite{chenteaching, olausson2023demystifying, gao2023rarr}. We comprehensively evaluate the performance of CRITIC \cite{gou2024critic}, which leverages specialized tools, such as code executors, to evaluate and reflect on the external feedback obtained through its interactions with these tools. However, recent studies \cite{huanglarge, stechly2023gpt, valmeekam2023can, zhang-etal-2024-self-contrast} cast doubt on the ability of LLMs to correct previous responses based solely on their inherent reflection capacity. Additionally, research on reflection with external feedback \cite{10.5555/3666122.3667845, shinn2023reflexion} have been doubted on using oracle labels which are are often unavailable in practice to determine when to stop the self-correction loop during the reflection. The improvements vanish when the reflection no longer relies on oracle labels \cite{huanglarge}. Apart from these studies questioning the use of oracle in self-reflection during the iterative reflection process, we also observe that improvements in \cite{gou2024critic} stem from using oracle labels to correct only the incorrect samples during evaluation.  Unlike previous findings, we observe that both intrinsic self-correction and external feedback can lead to a deterioration in performance after reflection. Additionally, our proposed Instruct of Reflection Tool (IoRT) diverges from existing reflection methods by emphasizing continuous performance optimization through diverse instructional guidance. We design the instructor to guide the next reflection iteration without relying on oracle labels.

\subsection{Abstract Reasoning}  
Instead of addressing the problems directly, abstract reasoning relies on higher-level thoughts to guide thinking \cite{zhengtake}, making it
essential for sophisticated cognitive processing in
artificial intelligence systems \cite{lake2017building, chollet2019measure, qiuphenomenal}. For example, consider the question "What are the roots of a quadratic equation?" Abstract reasoning involves extracting the fundamental principle of "the quadratic equation root formula" and applying this principle to this specific question. \cite{zhengtake} uses the concepts and principles to guide reasoning by taking a step back. \cite{fu2024hint} guides the model to generate hints such as specific knowledge or key ideas. \cite{xiong2024meaningful} designs a preliminary study to quantify and delve into the abstract reasoning abilities of LLMs. \cite{yang2024buffer} adapts by refining or retrieving high-level thought templates to instantiate specific reasoning structures. In our research, we build on \cite{fu2024hint} to generate meta-thoughts, thereby enhancing the abstract reasoning abilities of the reflective instructor. In contrast to the above works, which obtain an initial response through abstract reasoning, we incorporate abstract reasoning into the reflection process by designing the instructor within our framework.

\section{Empirical Evidence for Iterative Reflection }
\label{sec:Reflection Performance Evaluation}
\begin{figure}[t]
\centering
\includegraphics[width=1\linewidth]{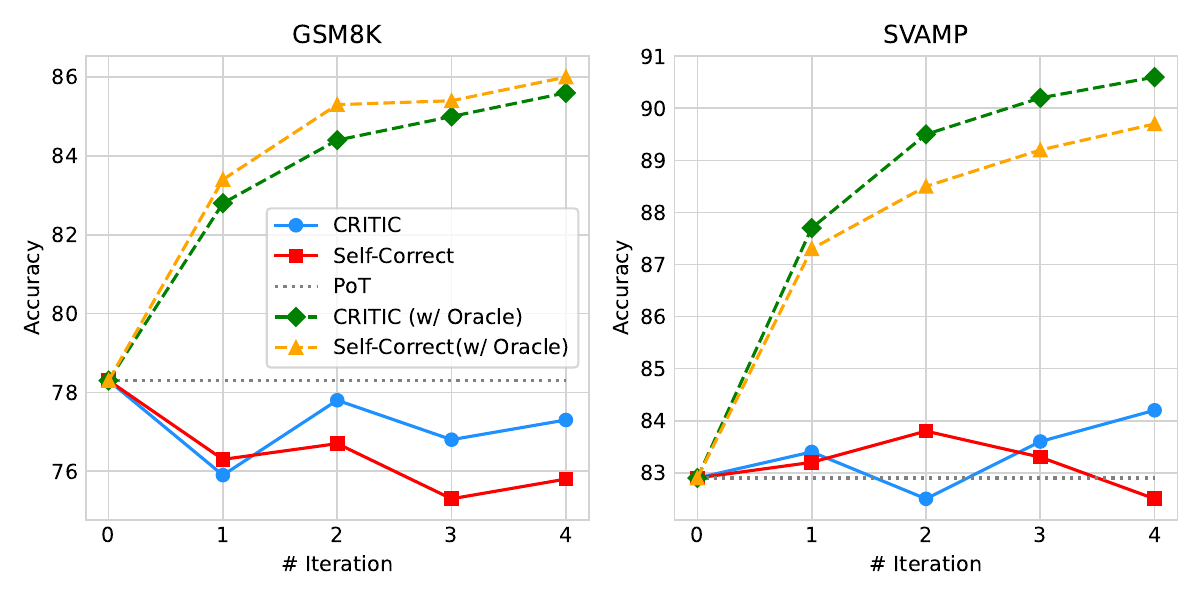}
\caption{Compare the performance changes of self-correct and CRITIC during iterative reflection across two datasets, evaluating with and without Oracle.}
\label{fig:2}
\end{figure}
Prior research primarily focuses on performance changes in a single reflection, which fails to fully capture the potential of reflection for long-term improvements. To gain deeper insights into how reflection can gradually refine answers, We comprehensively evaluate the performance changes of self-correct and CRITIC \cite{gou2024critic} across multiple reflection iterations \cite{huanglarge}.

\subsection{Performance of Iterative Reflection}
\label{subsec:Performance of Iterative Reflections}
We systematically evaluate the performance of self-correct (w/o external feedback) and CRITIC (w/ tool interactions as an available external feedback) across multiple iterative reflections on the GSM8K and SVAMP datasets with GPT-3.5. Additionally, we simulate an evaluation method (w/ oracle) that relies on oracle labels, focusing only on corrections from incorrect to correct answers in each iteration, while disregarding instances where correct answers are modified into incorrect ones.

In Figure~\ref{fig:2}, evaluating reflection using oracle labels, the accuracy of both self-correct and CRITIC improves steadily with each iteration. In contrast, without oracle labels, the performance is unstable and even deteriorate, regardless of the availability of external feedback. For instance, on GSM8K and SVAMP, the performances of self-correct and CRITIC drop by up to $-2.4\%$ and $-3.0\%$, respectively. Notably, compared to self-correct, the performance is more reliable with CRITIC.

\subsection{Further Performance Analysis}

Why does performance degrade after reflection? To investigate this question, we further conduct the following analysis:

\textbf{Step 1}: We classify all samples in GSM8K into four categories based on the correctness before and after reflection. Then analyze the performance changes in each iteration using self-correct and CRITIC on GPT-3.5. \textbf{Step 2}: We classify all samples into four categories based on the consistency and correctness of the answers throughout all iterations: 1) \textit{Redundant Iteration} means the answers consistently remain correct. 2) \textit{Invalid Consistent Iteration} insists on a consistent wrong answer. 3) \textit{Drift Iteration} generates both incorrect and correct during the iterative reflection. 4) \textit{Invalid Inconsistent Iteration} continuous changes occur in iterations without ever converging to the correct solution. The detailed results are as follows:
\begin{figure}[t]
\centering
\includegraphics[width=1\linewidth]{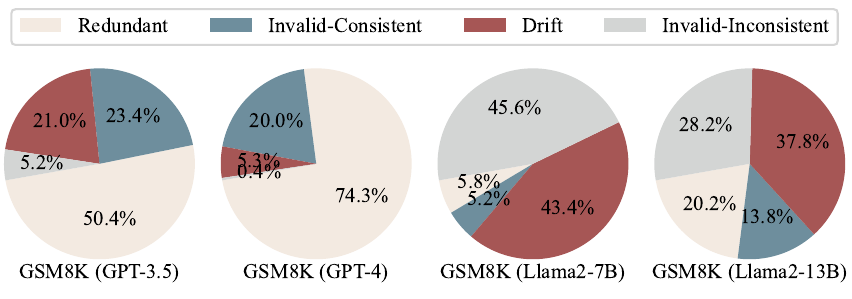}
\caption{Analyze the different types of iteration of CRITIC on GSM8K.}
\label{fig:3}
\end{figure}
\textbf{Reflection can not only be beneficial but also detrimental.} As shown in Table~\ref{table:critic}, for both self-correct and CRITIC, the proportion of \textcolor{green}{\ding{51}}$\Rightarrow$\textbf{\textcolor{red}{\ding{55}}} is often comparable to or even exceeds that of \textbf{\textcolor{red}{\ding{55}}}$\Rightarrow$\textcolor{green}{\ding{51}}, which explains the performance degradation after reflection. Notably, CRITIC demonstrates a lower ratio of \textcolor{green}{\ding{51}}$\Rightarrow$\textbf{\textcolor{red}{\ding{55}}} compared to self-correct, suggesting that external feedback promotes the stability of the reflection.

\textbf{Models of different sizes yield varying reflection performance.} In Figure~\ref{fig:3}, significant differences in reflection performance are observed across various LLMs. For instance, GPT-4 shows a $94.3\%$ share of stable iterations, while $89\%$ of LLaMA-2 7B’s iterations are unstable. It implies that larger models like GPT-4, benefiting from vast training data and substantial computational resources \cite{ouyang2022training, chowdhery2023palm, chung2024scaling}, exhibit more stable performance compared to smaller models.

\textbf{
Various reflective iterations correspond to different limitations.} As shown in Figure~\ref{fig:1}, LLMs often perform the $i_{\text{th}}$ reflection based on the $(i-1)_{\text{th}}$ response, and this static iterative reflection highlights the following three limitations: I. \textit{Redundant Iteration$\Rightarrow$ Redundant}:  Although redundant iterations do not affect accuracy, they contribute to unnecessary costs and delays. II. \textit{Drift Iteration$\Rightarrow$Drift}: Especially, \textcolor{green}{\ding{51}}$\Rightarrow$\textbf{\textcolor{red}{\ding{55}}} indicates that LLMs often exhibit a high degree of uncertainty, leading to reflection drift and a decline in accuracy. III. \textit{Invalid iteration$\Rightarrow$Stubborn}: LLMs are too stubborn to recognize and correct mistakes. Our method effectively addresses these three issues, achieving optimization of accuracy and efficiency in reflection during iterations.

\begin{figure*}[t]
\centering
\includegraphics[width=1\linewidth]{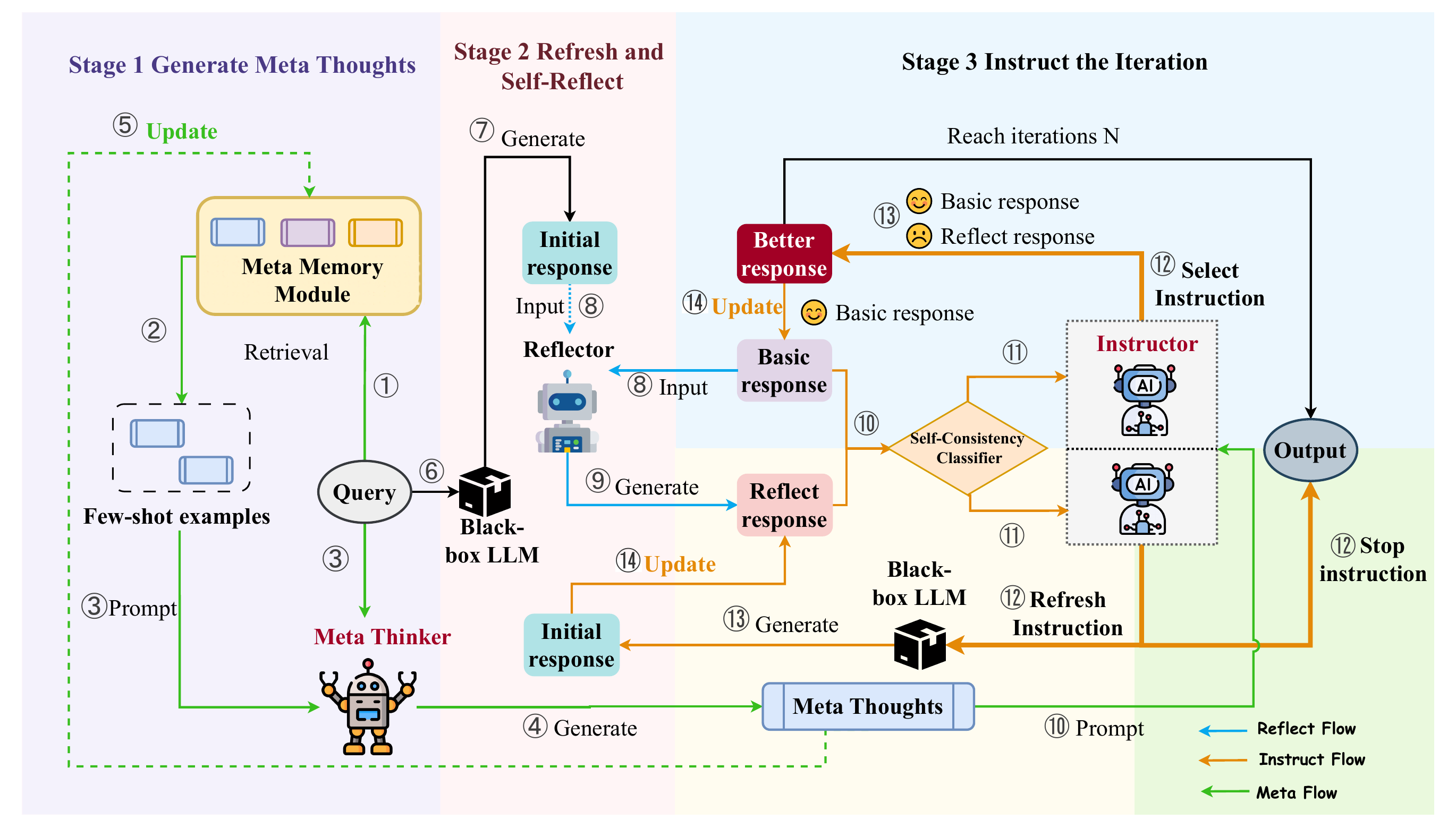}
\caption{IoRT comprises three steps: (1) generate meta-thoughts, (2) refresh and self-reflect, and (3) instruct the iteration. Repeat steps \normalsize{\textcircled{\scriptsize{8}}} to \normalsize{\textcircled{\scriptsize{14}}} after each iteration until the instructor signals to stop or the maximum iterations $N$ is reached.}
\label{fig:4}
\end{figure*}

\section{Instruct-of-Reflection (IoRT)}

Prior sections illustrate the challenges LLMs encounter in static iterative reflection such as redundancy, drift and stubborn. How to precisely identify effective reflections while mitigating the impact of harmful or redundant reflections? As a remedy, we innovatively propose Instruct-of-Reflection (IoRT), a novel framework that implements dynamic-meta instruction for the iterative reflection process. We can get an overview of the IoRT framework in Figure~\ref{fig:4}. Specifically, IoRT involves three main steps: generate meta-thoughts, refresh and self-reflect, and instruct the iteration. As the core module, the instructor is augmented by meta-thoughts and driven by self-consistency classifier, generating instructions including refresh, stop, and select, directing the next reflection iteration. 

\subsection{Generate Meta Thoughts}

Human often summarize and abstract higher-level ideas when solving problems \cite{yang2024buffer}. Inspired by Hint-before-Solving Prompting (HSP) \cite{fu2024hint} and Meta-Reasoning \cite{zeng2024mrgsm8kmetareasoningbenchmarklarge}, we propose meta-thinker which enables LLMs to explicitly generate meta-thoughts through a few-shot learning for problem solving. Meta-thoughts are high-level knowledge based on abstract reasoning, encompassing analytical methods and fundamental overall strategies. They enhance the instructor's role as a teacher by providing a comprehensive overview of the reflection process. For instance, for the question "Can a honey bee sting a human more than once?", since the bee's stinger is closely related to its ability to sting, the meta-thought for this question primarily focuses on the basic principle of the "changes in the stinger". The meta memory module stores meta-thoughts, distilled from various tasks, recorded in the form of $\mathcal{E} = \{(q_i, m_i)\}$,  expressing them as a pair relationship $e_i$ = ($q_i$, $m_i$), where $q_i$ represents the question statement and $m_i$ represents the meta-thought associated with solving question $q_i$.

Specifically, meta-thoughts construction includes 4 steps. \textbf{Initialization}: for each dataset, we randomly select $k$ questions and manually define their meta-thoughts as shown in Table~\ref{table:dataset} to initialize the meta-thought module. \textbf{Retrieval}: For the input question $x$, meta-thinker retrieves the k most relevant prompt examples by applying the cosine similarity function $S(q_i, x)$ to calculate the embedding similarity between the questions $q_i$ and $x$, the top $k$ questions and their corresponding meta-thoughts set are identified:
\begin{equation}
\{e^{1}_{\text{sim}}, e^{2}_{\text{sim}}, \dots, e^{k}_{\text{sim}}\} = \underset{q_i \in M}{\text{argmax}} \, S(q_i, x) \hspace{-10pt} 
\end{equation}
\begin{equation}
S(q_i, x) = \frac{\mathbf{q_i} \cdot \mathbf{x}}{\|\mathbf{q_i}\| \|\mathbf{x}\|}
\
\end{equation}
\textbf{Generation}: Based on these $k$ similar questions and their corresponding meta-thoughts, meta-thinker employs a model $f(\cdot)$ to generate meta-thought $m_x$ for the question $x$ through few-shot learning, which can be expressed as:
\begin{equation}
\hspace{-10pt} 
m_x = f\left(e^{1}_{\text{sim}}, e^{2}_{\text{sim}}, \dots, e^{k}_{\text{sim}}, x\right) \hspace{-10pt} 
\end{equation}
\textbf{Updating}: The new meta-thought $m_x$ not only enhances the instructor's abstract reasoning capabilities but also facilitates the updating of the meta-memory module, which continuously evolves by accumulating meta-thoughts for problem solving, thereby enabling more efficient and intelligent reasoning. In this context, the update of $\mathcal{E}$ can be formulated as:
\begin{equation}
\hspace{-10pt} 
\mathcal{E} \leftarrow \mathcal{E} \cup \{(x, m_x)\} \hspace{-10pt} 
\end{equation}
\subsection{Refresh and Self-Reflect}

In the refresh process, given a black-box LLM $g(\cdot)$ and a question $x$, the LLM utilizes its few-shot contextual learning ability to generate an initial output response $R^{0}_{o}$. In the $i_{\text{th}}$ iteration, we extracted the basic answer $A^{i}_{b}$ from the basic response $R^{i}_{b}$. The specifics of answer extraction from the response are elaborated in Appendix~\ref{sec:Datasets and Evaluation Metrics}. The reflector assesses the quality of $R^{i}_{b}$ using evaluation metrics such as plausibility and correctness, and provides feedback accordingly. Based on the input question $x$, the basic response $R^{i}_{b}$, the basic answer $A^{i}_{b}$ and the evaluation feedback $f_i$, a reflective response $R^{i}_{r}$ is generated as follows:
\begin{equation}
R^{i}_{r} = g(x, R^{i}_{b},A^{i}_{b}, f_i)
\end{equation}
\subsection{Instruct the Iteration}

As analyzed in (\S~\ref{sec:Reflection Performance Evaluation}), the answers after reflection may improve, degrade, or remain unchanged. Therefore, we design the instructor to manage the reflection process, ensuring stable performance improvement throughout the iterative reflection.

\textbf{Meta-Thought Augment Instructor} We incorporate the meta-thought $m_x$ into the instructor's prompt. During decision-making, the instructor treats the meta-thought as a critical evaluation criterion. By emphasizing abstract reasoning, the meta-thought elevates the LLMs from simply solving problems to instructing the process. This approach enables a more holistic evaluation of responses during each iteration, ensuring that the instructor's decisions not only satisfy problem-specific requirements but also undergo rigorous analysis.

\textbf{Self-Consistency Classifier} Does not use any LLM, self-consistency classifier determines consistency by comparing whether the basic answer $A^{i}_{b}$ and the reflective answer $A^{i}_{r}$ are equal. We implement a self-consistency classifier to evaluate the quality of responses, providing feedback to the instructor to guide decision-making. Based on the following three scenarios, the instructor generates a select, stop or refresh instruction:
\begin{itemize}[left=0pt] 

    \item \textbf{Select Instruction} If $A^{i}_{b} \neq A^{i}_{r}$, it suggests that at least one of the responses is incorrect. The instructor carefully evaluates both $R^{i}_{b}$ and $R^{i}_{r}$ based on the meta-thought $m_x$, and then selects the better response from them as $R^{i}_{o}$ represented as the output for the $i_{\text{th}}$ iteration. Concurrently, $R^{i+1}_{b}$ is updated to $R^{i}_{o}$, represented as:
\begin{equation}
R^{i}_{o} = \text{Instructor}(R^{i}_{b}, A^{i}_{b}, R^{i}_{r}, A^{i}_{r}, m_x, x)  
\end{equation}
If $A^{i}_{\text{b}}=A^{i}_{r}$, it indicates that the reflection did not change the output, we designate $R^{i}_{b}$ as  the output for the $i_{\text{th}}$ iteration $R^{i}_{o}$. The Instructor will evaluates both $R^{i}_{b}$ and $R^{i}_{r}$ to determine whether to issue a stop or refresh instruction:

\item \textbf{Stop Instruction} If the Instructor deems both $R^{i}_{b}$ and $R^{i}_{r}$ reasonable for solving the problem, a stop instruction is issued, and the iteration concludes at the $i_{\text{th}}$ round. In our experiments, we set the maximum number of iterations to $N$. To ensure comprehensive evaluation, all subsequent iterations are set equal to the response from the $i_{\text{th}}$ iteration, represented as:
\begin{equation}
R^{N}_{o}, R^{(N-1)}_{o}, \ldots, R^{(i+1)}_{o} \!= R^{i}_{o}
\end{equation}
\item \textbf{Refresh Instruction} 
If $R^{i}_{b}$ and $R^{i}_{r}$ fail to resolve the problem, in the $(i+1)_{\text{th}}$ iteration, a black-box LLM $g(\cdot)$ will generate a new response to update $R^{(i+1)}_{r}$. $R^{(i+1)}_{b}$ is set equal to $R^{i}_{b}$. IoRT will then compare $R^{(i+1)}_{b}$ and $R^{(i+1)}_{r}$ once again. Such invalid iterations can trap the model in a resource-intensive loop. By refreshing, we break this deadlock and introduce new ideas, enabling the model to generate more effective solutions.
\end{itemize}
\begin{table*}[t]
\Large
\centering
\renewcommand{\arraystretch}{1.2}  
\begin{adjustbox}{width=\textwidth}
\begin{tabular}{l*{11}{c}}  
\\\specialrule{0.07em}{0pt}{0pt}
\multirow{2}{*}{\centering \textbf{Methods}}  & \multicolumn{5}{c}{\textbf{GSM8K}} & \multicolumn{5}{c}{\textbf{SVAMP}} & \multirow{2}{*}{\centering \parbox{1.5cm}{\centering \textbf{\#Calls} \\ \textbf{Avg.}}}\\
                & \textbf{GPT-3.5} & \textbf{GPT-4} & \textbf{L-7B} & \textbf{L-13B} & \textbf{L-70B} & \textbf{GPT-3.5} & \textbf{GPT-4} & \textbf{L-7B} & \textbf{L-13B} & \textbf{L-70B} & \\
\hline
\multicolumn{12}{c}{\parbox[c][1.2em][c]{\textwidth}{\centering \textit{Text Describing Reasoning}}} \\
\hline

CoT              & 76.6 & 93.9 & 19.8 & 28.3 & 52.6 & 79.8 & 93.0 & 37.5 & 40.2 & 66.0 & 1 \\
PS               & 75.7 & 94.1 & 20.3 & 30.4 & 57.8 & 83.6 & 93.3 & 45.2 & 53.7 & 70.9 & 1 \\
CoT+HSP          & 80.1 & 94.0 & 20.7 & 32.3 & 58.9 & 83.7 & 92.9 & 40.4 & 50.3 & 72.5 & 2 \\
Self-Contrast    & 84.4 & 95.4 & 20.5 & \textbf{42.3} & 64.2 & \textbf{89.0} & 94.0 & 44.5 & 54.6 & 75.3 & 7.8 \\
CoT-SC(8)           & 83.5 & 94.2 & 21.4 & 37.6 & 61.1 & 84.6 & 92.5 & 45.2 & 53.7 & 72.0 & 8 \\
Multi-Agent      & 83.8 & 93.5 & 23.8 & 34.9 & 59.6 & 84.1 & 93.2 & 42.5 & 49.2 & 70.1 & 9 \\
Self-Reflection   & 75.8 & 95.1 & 17.0 & 31.8 & 49.3 & 80.5 & 91.5 & 36.1 & 42.5 & 63.0 & 9 \\
\hline
\multicolumn{12}{c}{\parbox[c][1.2em][c]{\textwidth}{\centering \textit{Programming Language Reasoning}}} \\
\hline
PoT & 78.3 & 94.0 & 19.5 & 30.5 & 58.6 & 82.9 & 93.4 & 43.5 & 53.0 & 75.2 & 1 \\
PoT+HSP          & 84.0 & 94.9 & 17.0 & 35.1 & 64.9 & 86.3 & 94.7 & 43.6 & 55.1 & 78.4 & 2 \\
PoT-SC(8)              & 83.7 & 94.4 & 22.5 & 37.7 & 64.3 & 87.0 & 93.8 & 45.3 & 56.0 & 77.8 & 8 \\

Self-Reflection & 75.8 & 94.5 & 15.7 & 25.6 & 55.5 & 82.5 & 90.1 & 33.8 & 47.9 & 70.7 & 9 \\
CRITIC           & 77.3 & 93.7 & 14.3 & 26.8 & 56.7 & 84.2 & 93.9 & 38.2 & 45.6 & 72.0 & 9 \\
\hline
IoRT       & \textbf{84.6} & \textbf{95.4} & \textbf{24.0} & 40.8 & \textbf{66.0} & 88.1 & \textbf{95.3} & \textbf{45.7} & \textbf{57.5} & \textbf{80.1} & 7.3 
\\\specialrule{0.07em}{0pt}{0pt}
\end{tabular}
\end{adjustbox}
\caption{The performance of IoRT on mathematical reasoning. IoRT employs PoT to generate the initial responses. The best results are highlighted in bold. The text describing reasoning experimental results for CoT, CoT-SC(8), Multi-Agent, Self-Contrast, and Self-Reflection were sourced from \cite{zhang-etal-2024-self-contrast}. L- denotes Llama2-chat.}
\label{tab:2}
\end{table*}

\begin{table}[htbp]
\centering
\Large
\renewcommand{\arraystretch}{1.3} 
\resizebox{\linewidth}{!}{ 
\begin{tabular}{lcccccc}
\hline
\textbf{Methods}   & \textbf{GPT-3.5} & \textbf{GPT-4} & \textbf{L-7B} & \textbf{L-13B} & \textbf{L-70B} & \textbf{\#Token Num.}\\
\hline
CoT               & 66.8 & 75.4 & 47.3 & 54.2 & 62.0 & 514\\
CoT-SC(8)            & 69.1 & 77.0 & 49.4 & 57.1 & 63.6 & 4145 \\
CoT+HSP           & 71.5    & 77.9    & 50.5    & 56.3    & 65.2 & 1018 \\
PS               & 68.2 & 76.1 & 48.3 & 57.5 & 64.9 & 1090\\
Multi-Agent       & 68.5 & 76.6 & 50.8 & 56.8 & 65.7 & 3922 \\
Self-Reflection   & 60.7 & 76.9 & 36.8 & 47.0 & 54.3 & 5944 \\
\hline
IoRT              & \textbf{71.9} & \textbf{78.8} & \textbf{53.1}    & \textbf{60.1} & \textbf{67.9} & 3877\\
\hline
\end{tabular}
}
\caption{The performance in the final iteration on commonsense reasoning. We utilize CoT to produce the initial responses. For the Multi-Agent, we configure three agents to engage in a three-round debate.}
\label{tab:3}
\end{table}
\section{Experiments}

\subsection{Experimental Setup}

\textbf{Benchmark} We evaluated our method in two reasoning scenarios: mathematical reasoning and commonsense reasoning. Specifically, for commonsense reasoning, we use the StrategyQA \cite{geva-etal-2021-aristotle}, while for mathematical reasoning, we use GSM8K \cite{cobbe2021training} and SVAMP \cite{patel-etal-2021-nlp}. These three datasets are widely adopted in existing research, serving as standard benchmarks for evaluating reasoning capabilities.

\textbf{Baselines} We compare IoRT with the following strong baselines: 1. Chain-of-Thought prompting (CoT) \cite{wei2022chain}; 2. Self-Consistency (SC) \cite{wangself}; 3. Plan-and-Solve Prompting (PS) \cite{wang-etal-2023-plan}; 4. Multi-Agent Debate \cite{du2023improving,liang2023encouraging}; 5. Self-Contrast \cite{zhang-etal-2024-self-contrast}; 6. Program-of-thought (PoT) \cite{chen2022program}; 7. Hint-before-Solving Prompting \cite{fu2024hint}; 8. Self-Reflection \cite{shinn2023reflexion}; 9. Self-Correcting with Tool-Interactive Critiquing (CRITIC) \cite{gou2024critic}. For simplicity in notation, we use "CoT-SC(8)" and "PoT-SC(8)" to denote the approach that retrieves eight CoT or PoT reasoning chains to make majority vote. 

\textbf{Implementation Details} In Figure~\ref{fig:4}, we use the GPT-3.5-Turbo-0613, GPT-4-0613, and Llama2-Chat models at three parameter scales (7B, 13B, and 70B) for black-box LLM and the reflector. Throughout the experiment, GPT-3.5-Turbo-0613 serves as both the meta-thinker and the instructor. We uniformly set the temperature to 0.3 and the maximum number of iterations to 4. To ensure the reliability of the results, we conduct five rounds of experiments for each dataset and report the average scores as the final evaluation results. For evaluation metrics, we report accuracy. To evaluate the computational cost, we also report the average number of API/LLM calls (\#Calls Avgs.) \cite{zhang-etal-2024-self-contrast} across three datasets,  as well as the average number of tokens consumed per question (\#Tokens Num.) for commonsense reasoning tasks.

\subsection{Main Results}

\textbf{Mathematical Reasoning.} According to the results presented in Table~\ref{tab:2}, IoRT demonstrates superior performance improvement compared to the current strong baselines. Our method achieves an average improvement of f approximately $+4.4\%$ compared to PoT. In contrast, CRITIC, which also generate initial responses based on PoT, results in performance decreases of $-2.6\%$. Besides, compared to the text describing reasoning baselines such as CoT, our method also achieve significant improvements of $+8.9\%$. It is worth noting that IoRT achieves an average performance improvement of $+2.4\%$ compared to PoT+HSP directly incorporating abstract reasoning to generate an initial response. This indicates that integrating meta-thought into the instructor optimizes reflecting continuously. Moreover, IoRT reduces average call overhead by approximately 27.6\% compared to iterative reflection, demonstrating that our approach effectively balances accuracy and call overhead. Although self-contrast outperforms our method on GSM8K with $Llama2_{7B}$ and on SVAMP with GPT-3.5, this improvement comes at the cost of increased call overheads. Specifically, self-contrast incurs 7.0\% higher overhead compared to IoRT. 

\textbf{Commonsense Reasoning. } Table~\ref{tab:3} compares the performance of IoRT on commonsense reasoning tasks. We observe that IoRT outperforms the baseline across all models on StrategyQA. Specifically, our method gains $+5.2\%$ performance increase over CoT, whereas self-reflection results in a decrease of $-6.0\%$. IoRT also achieves an average performance improvement of $2.1\%$ compared to CoT+HSP on StrategyQA, demonstrating the efficacy of incorporating meta-thought into the reflection process. Additionally, with a token consumption  of 3877, IoRT significantly reduces overhead compared to high-cost methods like self-reflection (5944) and CoT-SC(8) (4145) while maintaining competitive reasoning performance.

\begin{table}[h]
\centering
\Large
\renewcommand{\arraystretch}{1.2} 
\setlength{\tabcolsep}{4pt} 
\scriptsize 
\begin{adjustbox}{width=\linewidth}
\begin{tabular}{llccccc}
\hline
\multirow{2}{*}{\textbf{Model}}   & \multirow{2}{*}{\textbf{Datasets}} & \multirow{2}{*}{\textbf{Initial}}   & \multirow{2}{*}{\textbf{IoRT*}} &\textbf{IoRT} &\textbf{IoRT}& \multirow{2}{*}{\textbf{IoRT}}\\ 
                 & & &      & (w/o SC)&(w/o MT)\\ 
\hline
\multirow{3}{*}{GPT-3.5} & GMS8K  & 78.3 & 80.6 & 84.4 & 83.0 & 84.6 \\
                                  & SVAMP  & 82.9  & 83.8 & 89.0 & 86.4 & 88.1 \\
                                  & StrategyQA    & 66.8  & 65.9 & 71.6 & 70.5 & 71.9 \\
\hline
\multirow{3}{*}{GPT-4}   & GMS8K  & 94.0 & 94.1 & 94.6 & 94.3 & 95.4 \\
                                  & SVAMP  & 93.4  & 93.8 & 93.9 & 93.7 & 95.3 \\
                                  & StrategyQA    & 75.4  & 77.0 & 77.7 & 76.6 & 78.8 \\
\hline
\multirow{3}{*}{L-7B} & GMS8K & 19.5  & 16.3 & 23.5 & 21.5 & 24.0 \\
                                  & SVAMP  & 43.5 & 43.4 & 45.1 & 44.2 & 45.7 \\
                                  & StrategyQA    & 47.3 & 47.7 & 51.3 & 49.7 & 53.1 \\
\hline
\multirow{3}{*}{L-13B} & GMS8K & 30.5  & 33.2 & 39.5 & 36.6 & 40.8 \\
                                   & SVAMP  & 53.0  & 53.7 & 56.9 & 55.4 & 57.5 \\
                                   & StrategyQA    & 54.2  & 52.2 & 60.5 & 58.7 & 60.1 \\
\hline
\multirow{3}{*}{L-70B} & GMS8K & 58.6  & 61.8 & 66.3 & 64.1 & 66.0 \\
                                   & SVAMP  & 75.2  & 76.1 & 78.8 & 77.9 & 80.1 \\
                                   & StrategyQA   & 62.0 & 63.2 & 68.5 & 65.6 & 67.9 \\
\hline     
\end{tabular}
\end{adjustbox}
\caption{Performance comparisons upon different components and settings under instructing 
reflection strategy. Evaluation on the final iteration. Initial indicates the performance of generating the initial response.}
\label{table:ablation_study}
\end{table}

\subsection{Ablation Study}
We perform an ablation study to assess the impact of each steps on model performance:

\begin{itemize}[left=0pt]

\item IoRT*: This setting indicates the result of the $i_{\text{th}}$ iteration $R^{i}_{o}$ is modified as $R^{i}_{r}$.

\item IoRT(w/o SC): It indicates that regardless of whether the responses $R^{i}_{b}$ and $R^{i}_{r}$ are consistent, the instructor only selects until reaches the maximum number of iterations $N$.

\item IoRT (w/o MT): It implies that meta-thoughts are not provided and instructor relies solely on its own capacity to generate instructions.
\end{itemize}

As shown in Table~\ref{table:ablation_study}, we find that modules and settings of IoRT positively impact accuracy and efficiency improvements. 

\textbf{Select instruction effectively mitigates drift in reflections.} IoRT* exhibits a performance decrease of -4.4\% compared to IoRT, emphasizing that LLMs frequently struggle with generation stability during reflection. Thus, selecting instruction is essential for successful iterative reflection.

\textbf{Self-consistency has minimal impact on accuracy but significantly improves efficiency.}  IoRT(w/o SC) results in no clear change(-0.51\%) in accuracy compared to IoRT, indicating that the instructor enables accurately identifying the better response during the iterations. However, in the experiment, IoRT (w/o SC) requires four iterations to stop, while the average number of iterations for IoRT is just 2.2, indicating that self-consistency can reduce overhead.

\textbf{Meta-thought enhances the reasoning capabilities of the Instructor.} IoRT (w/o MT) shows a $-2.1\%$ decline compared to IoRT but still achieves a $+2.9\%$ improvement over the initial response. This highlights the positive impact of our dynamic iterative framework on reflection, with meta-thoughts further enhancing the instructor's ability to accurately identify correct reflections.

\section{Discussions}
\begin{figure}[t]
\centering
\includegraphics[width=1\linewidth]{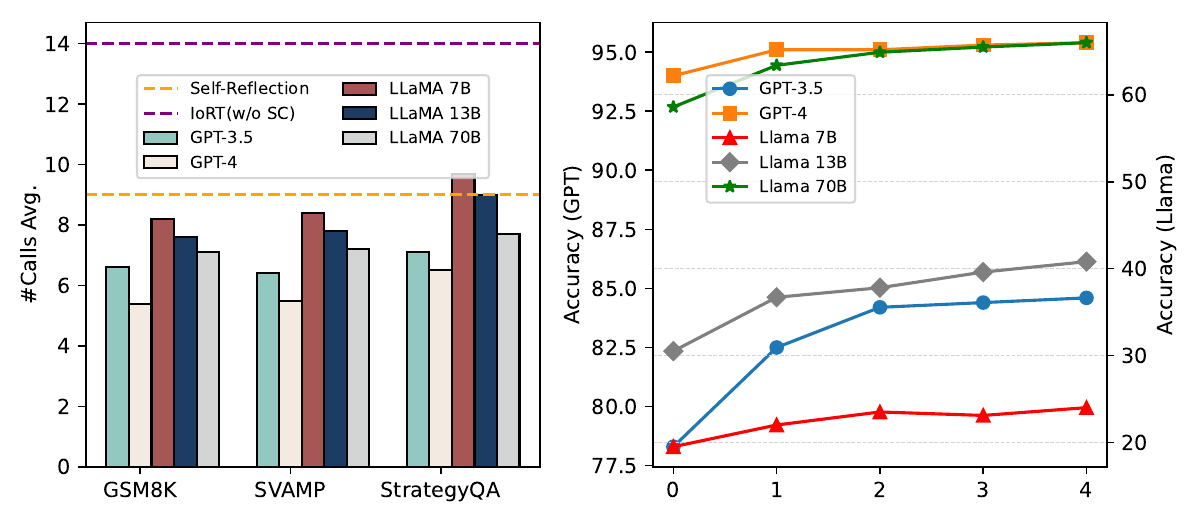}
\caption{Left: Comparison of the average number of API/LLM calls across various datasets on IoRT. Right: The performance of GPT and Llama-2 models on GSM8K with IoRT. }
\label{fig:5}
\end{figure}
\textbf{Performance Analysis.} In \S~\ref{sec:Reflection Performance Evaluation}, our experiments indicate that the performance of CRITIC even deteriorates. In Figure~\ref{fig:5} (right), IoRT achieves stable improvements across different LLMs. This is because CRITIC follows a static reflection pipeline: initial response → evaluation → revision. In contrast, IoRT adopts a dynamic iterative pipeline: initial response → evaluation → revision → instruction. By comparing Table~\ref{table:critic} and Table~\ref{table:IoRT}, we observe that with each iteration of IoRT, the occurrences of \textcolor{green}{\ding{51}}$\Rightarrow$\textbf{\textcolor{red}{\ding{55}}} and \textbf{\textcolor{red}{\ding{55}}}$\Rightarrow$\textbf{\textcolor{red}{\ding{55}}} cases significantly reduced. The results indicate that IoRT can identify \textcolor{green}{\ding{51}} in drifting iterations through select instructions, or generate a refresh instruction to promptly terminate stubborn iterations, effectively ensuring stable improvement and optimization.

\textbf{Overheads and Manual Efforts Analysis.} In Figure~\ref{fig:5} (left), we investigate the average number of API/LLM calls across various datasets on IoRT. In our experiments, we set the maximum number of iterations to 4. Self-reflection and IoRT (w/o SC) were configured with a fixed number of iterations, leading to 9 and 14 call overheads, respectively. Compared to self-reflection and IoRT (w/o SC), IoRT achieves substantial improvements, reducing call overheads by more than $18.8\%$ and $47.9\%$, respectively. Rather than simply halting reflection after the maximum number of iterations, IoRT integrates self-consistency checks and meta-thoughts, enabling the identification of correct responses and the generation of stop instructions to timely halt iterations,  significantly reducing redundancy and overhead. Moreover, our method defines only three roles: meta-thinker, reflector, and instructor, which is considerably fewer than multi-agent debate. This simplification eases the manual effort required for pre-configuring agent roles.

\textbf{Model Generality and Differences.} As shown in Table~\ref{tab:2}, compared to CoT-SC(8), IoRT achieves significant performance improvements of $2.5\%$ on GPT-3.5, $1.9\%$ on GPT-4, $2.3\%$ on $Llama2_{7B}$, $3.3\%$ on $Llama2_{13B}$, and $5.8\%$ on $Llama2_{70B}$, with an average improvement across the three datasets, indicating that IoRT demonstrates robust generality across both commercial LLMs (GPT) and open-source models (Llama-2).
Moreover, these results indicate that IoRT often achieves more significant improvements with smaller-scale models, indicating its effectiveness in optimizing the performance of models with fewer parameters. This could be because larger-scale models produce more stable outputs, while smaller-scale models are more likely to generate varied responses through reflection, resulting in greater diversity in their outputs. In prior studies  relying on static reflection methods, performance becomes unstable due to the influence of such diversity, and the ratio of \textcolor{green}{\ding{51}}$\Rightarrow$\textbf{\textcolor{red}{\ding{55}}} can lead to performance degradation. However, IoRT can quickly capture correct answers  from diverse outputs, leading to notable performance improvements. The key difference enabling this is the select instruction in IoRT, which is unaffected by \textcolor{green}{\ding{51}}$\Rightarrow$\textbf{\textcolor{red}{\ding{55}}} and ensures stable performance improvement due to \textcolor{red}{\ding{55}}$\Rightarrow$\textbf{\textcolor{green}{\ding{51}}}. Notably, $Llama2_{70B}$ demonstrates a greater ability to generate accurate responses through reflection compared to $Llama2_{7B}$ and $Llama2_{13B}$. With the smaller parameter sizes, $Llama2_{7B}$ and $Llama2_{13B}$ struggle to produce correct answers, leading to more frequent invalid and inconsistent iterations. 

Analysis of Figure~\ref{fig:5} (left) shows differences in call overheads for IoRT across various models. GPT averages 1.9 fewer LLM calls than Llama2, possibly because larger-scale models tend to produce more redundant iterations, so they often stop early with self-consistency checks. In contrast, smaller-scale models have less stable outputs, resulting in more iterations.

\section{Conclusion}

We conducted a comprehensive analysis of the iterative reflection performance of LLMs. The empirical evidence suggests that the performances of these reflection methods are unsatisfactory, primarily due to the limitations of static iterative reflection, which leads to redundant, drift, and stubborn issues. To mitigate this, we propose Instruct-of-Reflection (IoRT), a dynamic iterative reflection framework that integrates abstract reasoning into the reflection,  generating adaptive instruction to regulate the iterative reflection. Experimental results demonstrate that IoRT surpasses a series of strong baselines and reduces overhead across
a variety of scenarios and with different LLMs.


\section*{Limitations}

Despite the significant improvements of our method over a series of strong baselines, it is not always able to provide perfect guidance during the reflection process. For instance, as shown in Figure~\ref{fig:2}, in experiments on mathematical reasoning with GPT-3.5, the final iteration performance, when evaluated using oracle labels, exceeds our method by 1.6\%. This indicates that occasional misjudgments still occur in IoRT, though their frequency remains relatively low. Additionally, due to the current limitations of open-source models \cite{touvron2023llama} in abstract reasoning and guiding the reflection process, we did not use them as our meta-thinker and instructor in this experiment. In future work, we try to address these limitations by exploring fine-tuning, distillation, or other methods to enhance open-source models \cite{achiam2023gpt}, equipping them with robust abstract reasoning and instructional capabilities, potentially enabling them to match or even exceed the performance of commercial models.
\section*{Acknowledgements}
We appreciate the reviewers for their insightful
comments and suggestions. This work was supported by the Beijing Association of Higher Education project MS2023151.

\bibliography{custom}

\begin{thebibliography}{50}
\providecommand{\natexlab}[1]{#1}

\bibitem[{Achiam et~al.(2023)Achiam, Adler, Agarwal, Ahmad, Akkaya, Aleman, Almeida, Altenschmidt, Altman, Anadkat et~al.}]{achiam2023gpt}
Josh Achiam, Steven Adler, Sandhini Agarwal, Lama Ahmad, Ilge Akkaya, Florencia~Leoni Aleman, Diogo Almeida, Janko Altenschmidt, Sam Altman, Shyamal Anadkat, et~al. 2023.
\newblock Gpt-4 technical report.
\newblock \emph{arXiv preprint arXiv:2303.08774}.

\bibitem[{Anil et~al.(2023)Anil, Dai, Firat, Johnson, Lepikhin, Passos, Shakeri, Taropa, Bailey, Chen et~al.}]{anil2023palm}
Rohan Anil, Andrew~M Dai, Orhan Firat, Melvin Johnson, Dmitry Lepikhin, Alexandre Passos, Siamak Shakeri, Emanuel Taropa, Paige Bailey, Zhifeng Chen, et~al. 2023.
\newblock Palm 2 technical report.
\newblock \emph{arXiv preprint arXiv:2305.10403}.

\bibitem[{Bai et~al.(2022{\natexlab{a}})Bai, Jones, Ndousse, Askell, Chen, DasSarma, Drain, Fort, Ganguli, Henighan et~al.}]{bai2022training}
Yuntao Bai, Andy Jones, Kamal Ndousse, Amanda Askell, Anna Chen, Nova DasSarma, Dawn Drain, Stanislav Fort, Deep Ganguli, Tom Henighan, et~al. 2022{\natexlab{a}}.
\newblock Training a helpful and harmless assistant with reinforcement learning from human feedback.
\newblock \emph{arXiv preprint arXiv:2204.05862}.

\bibitem[{Bai et~al.(2022{\natexlab{b}})Bai, Kadavath, Kundu, Askell, Kernion, Jones, Chen, Goldie, Mirhoseini, McKinnon, Chen, Olsson, Olah, and et~al.}]{bai2022constitutionalaiharmlessnessai}
Yuntao Bai, Saurav Kadavath, Sandipan Kundu, Amanda Askell, Jackson Kernion, Andy Jones, Anna Chen, Anna Goldie, Azalia Mirhoseini, Cameron McKinnon, Carol Chen, Catherine Olsson, Christopher Olah, and et~al. 2022{\natexlab{b}}.
\newblock \href {https://arxiv.org/abs/2212.08073} {Constitutional ai: Harmlessness from ai feedback}.
\newblock \emph{Preprint}, arXiv:2212.08073.

\bibitem[{Chen et~al.(2023)Chen, Ma, Wang, and Cohen}]{chen2022program}
Wenhu Chen, Xueguang Ma, Xinyi Wang, and William~W. Cohen. 2023.
\newblock Program of thoughts prompting: Disentangling computation from reasoning for numerical reasoning tasks.
\newblock \emph{Transactions on Machine Learning Research}.

\bibitem[{Chen et~al.(2024)Chen, Lin, Sch{\"a}rli, and Zhou}]{chenteaching}
Xinyun Chen, Maxwell Lin, Nathanael Sch{\"a}rli, and Denny Zhou. 2024.
\newblock Teaching large language models to self-debug.
\newblock In \emph{The Twelfth International Conference on Learning Representations}.

\bibitem[{Chollet(2019)}]{chollet2019measure}
Fran{\c{c}}ois Chollet. 2019.
\newblock On the measure of intelligence.
\newblock \emph{arXiv preprint arXiv:1911.01547}.

\bibitem[{Chowdhery et~al.(2023)Chowdhery, Narang, Devlin, Bosma, Mishra, Roberts, Barham, Chung, Sutton, Gehrmann et~al.}]{chowdhery2023palm}
Aakanksha Chowdhery, Sharan Narang, Jacob Devlin, Maarten Bosma, Gaurav Mishra, Adam Roberts, Paul Barham, Hyung~Won Chung, Charles Sutton, Sebastian Gehrmann, et~al. 2023.
\newblock Palm: Scaling language modeling with pathways.
\newblock \emph{Journal of Machine Learning Research}, 24(240):1--113.

\bibitem[{Chu et~al.(2023)Chu, Chen, Chen, Yu, He, Wang, Peng, Liu, Qin, and Liu}]{chu2023survey}
Zheng Chu, Jingchang Chen, Qianglong Chen, Weijiang Yu, Tao He, Haotian Wang, Weihua Peng, Ming Liu, Bing Qin, and Ting Liu. 2023.
\newblock A survey of chain of thought reasoning: Advances, frontiers and future.
\newblock \emph{arXiv preprint arXiv:2309.15402}.

\bibitem[{Chung et~al.(2024)Chung, Hou, Longpre, Zoph, Tay, Fedus, Li, Wang, Dehghani, Brahma et~al.}]{chung2024scaling}
Hyung~Won Chung, Le~Hou, Shayne Longpre, Barret Zoph, Yi~Tay, William Fedus, Yunxuan Li, Xuezhi Wang, Mostafa Dehghani, Siddhartha Brahma, et~al. 2024.
\newblock Scaling instruction-finetuned language models.
\newblock \emph{Journal of Machine Learning Research}, 25(70):1--53.

\bibitem[{Claude(2024)}]{TheC3}
Claude. 2024.
\newblock \href {https://api.semanticscholar.org/CorpusID:268232499} {The claude 3 model family: Opus, sonnet, haiku}.

\bibitem[{Cobbe et~al.(2021)Cobbe, Kosaraju, Bavarian, Chen, Jun, Kaiser, Plappert, Tworek, Hilton, Nakano et~al.}]{cobbe2021training}
Karl Cobbe, Vineet Kosaraju, Mohammad Bavarian, Mark Chen, Heewoo Jun, Lukasz Kaiser, Matthias Plappert, Jerry Tworek, Jacob Hilton, Reiichiro Nakano, et~al. 2021.
\newblock Training verifiers to solve math word problems.
\newblock \emph{arXiv preprint arXiv:2110.14168}.

\bibitem[{Du et~al.(2023)Du, Li, Torralba, Tenenbaum, and Mordatch}]{du2023improving}
Yilun Du, Shuang Li, Antonio Torralba, Joshua~B Tenenbaum, and Igor Mordatch. 2023.
\newblock Improving factuality and reasoning in language models through multiagent debate.
\newblock \emph{arXiv preprint arXiv:2305.14325}.

\bibitem[{Fu et~al.(2024)Fu, Huangfu, Yan, Ng, and Qiu}]{fu2024hint}
Jinlan Fu, Shenzhen Huangfu, Hang Yan, See-Kiong Ng, and Xipeng Qiu. 2024.
\newblock Hint-before-solving prompting: Guiding llms to effectively utilize encoded knowledge.
\newblock \emph{arXiv preprint arXiv:2402.14310}.

\bibitem[{Ganguli et~al.(2023)Ganguli, Askell, Schiefer, Liao, Luko{\v{s}}i{\=u}t{\.e}, Chen, Goldie, Mirhoseini, Olsson, Hernandez et~al.}]{ganguli2023capacity}
Deep Ganguli, Amanda Askell, Nicholas Schiefer, Thomas~I Liao, Kamil{\.e} Luko{\v{s}}i{\=u}t{\.e}, Anna Chen, Anna Goldie, Azalia Mirhoseini, Catherine Olsson, Danny Hernandez, et~al. 2023.
\newblock The capacity for moral self-correction in large language models.
\newblock \emph{arXiv preprint arXiv:2302.07459}.

\bibitem[{Gao et~al.(2023)Gao, Dai, Pasupat, Chen, Chaganty, Fan, Zhao, Lao, Lee, Juan et~al.}]{gao2023rarr}
Luyu Gao, Zhuyun Dai, Panupong Pasupat, Anthony Chen, Arun~Tejasvi Chaganty, Yicheng Fan, Vincent Zhao, Ni~Lao, Hongrae Lee, Da-Cheng Juan, et~al. 2023.
\newblock Rarr: Researching and revising what language models say, using language models.
\newblock In \emph{Proceedings of the 61st Annual Meeting of the Association for Computational Linguistics (Volume 1: Long Papers)}, pages 16477--16508.

\bibitem[{Geva et~al.(2021)Geva, Khashabi, Segal, Khot, Roth, and Berant}]{geva-etal-2021-aristotle}
Mor Geva, Daniel Khashabi, Elad Segal, Tushar Khot, Dan Roth, and Jonathan Berant. 2021.
\newblock \href {https://doi.org/10.1162/tacl_a_00370} {Did aristotle use a laptop? a question answering benchmark with implicit reasoning strategies}.
\newblock \emph{Transactions of the Association for Computational Linguistics}, 9:346--361.

\bibitem[{Gou et~al.(2024)Gou, Shao, Gong, yelong shen, Yang, Duan, and Chen}]{gou2024critic}
Zhibin Gou, Zhihong Shao, Yeyun Gong, yelong shen, Yujiu Yang, Nan Duan, and Weizhu Chen. 2024.
\newblock \href {https://openreview.net/forum?id=Sx038qxjek} {{CRITIC}: Large language models can self-correct with tool-interactive critiquing}.
\newblock In \emph{The Twelfth International Conference on Learning Representations}.

\bibitem[{Huang et~al.(2024)Huang, Chen, Mishra, Zheng, Yu, Song, and Zhou}]{huanglarge}
Jie Huang, Xinyun Chen, Swaroop Mishra, Huaixiu~Steven Zheng, Adams~Wei Yu, Xinying Song, and Denny Zhou. 2024.
\newblock Large language models cannot self-correct reasoning yet.
\newblock In \emph{The Twelfth International Conference on Learning Representations}.

\bibitem[{Kim et~al.(2024{\natexlab{a}})Kim, Baldi, and McAleer}]{kim2024language}
Geunwoo Kim, Pierre Baldi, and Stephen McAleer. 2024{\natexlab{a}}.
\newblock Language models can solve computer tasks.
\newblock \emph{Advances in Neural Information Processing Systems}, 36.

\bibitem[{Kim et~al.(2024{\natexlab{b}})Kim, Baldi, and McAleer}]{10.5555/3666122.3667845}
Geunwoo Kim, Pierre Baldi, and Stephen McAleer. 2024{\natexlab{b}}.
\newblock Language models can solve computer tasks.
\newblock In \emph{Proceedings of the 37th International Conference on Neural Information Processing Systems}, NIPS '23, Red Hook, NY, USA. Curran Associates Inc.

\bibitem[{Lake et~al.(2017)Lake, Ullman, Tenenbaum, and Gershman}]{lake2017building}
Brenden~M Lake, Tomer~D Ullman, Joshua~B Tenenbaum, and Samuel~J Gershman. 2017.
\newblock Building machines that learn and think like people.
\newblock \emph{Behavioral and brain sciences}, 40:e253.

\bibitem[{Liang et~al.(2023)Liang, He, Jiao, Wang, Wang, Wang, Yang, Tu, and Shi}]{liang2023encouraging}
Tian Liang, Zhiwei He, Wenxiang Jiao, Xing Wang, Yan Wang, Rui Wang, Yujiu Yang, Zhaopeng Tu, and Shuming Shi. 2023.
\newblock Encouraging divergent thinking in large language models through multi-agent debate.
\newblock \emph{arXiv preprint arXiv:2305.19118}.

\bibitem[{Madaan et~al.(2024)Madaan, Tandon, Gupta, Hallinan, Gao, Wiegreffe, Alon, Dziri, Prabhumoye, Yang, Gupta, Majumder, Hermann, Welleck, Yazdanbakhsh, and Clark}]{10.5555/3666122.3668141}
Aman Madaan, Niket Tandon, Prakhar Gupta, Skyler Hallinan, Luyu Gao, Sarah Wiegreffe, Uri Alon, Nouha Dziri, Shrimai Prabhumoye, Yiming Yang, Shashank Gupta, Bodhisattwa~Prasad Majumder, Katherine Hermann, Sean Welleck, Amir Yazdanbakhsh, and Peter Clark. 2024.
\newblock Self-refine: iterative refinement with self-feedback.
\newblock In \emph{Proceedings of the 37th International Conference on Neural Information Processing Systems}, NIPS '23, Red Hook, NY, USA. Curran Associates Inc.

\bibitem[{Olausson et~al.(2023)Olausson, Inala, Wang, Gao, and Solar-Lezama}]{olausson2023demystifying}
Theo~X Olausson, Jeevana~Priya Inala, Chenglong Wang, Jianfeng Gao, and Armando Solar-Lezama. 2023.
\newblock Demystifying gpt self-repair for code generation.
\newblock \emph{arXiv preprint arXiv:2306.09896}.

\bibitem[{Ouyang et~al.(2022)Ouyang, Wu, Jiang, Almeida, Wainwright, Mishkin, Zhang, Agarwal, Slama, Ray et~al.}]{ouyang2022training}
Long Ouyang, Jeffrey Wu, Xu~Jiang, Diogo Almeida, Carroll Wainwright, Pamela Mishkin, Chong Zhang, Sandhini Agarwal, Katarina Slama, Alex Ray, et~al. 2022.
\newblock Training language models to follow instructions with human feedback.
\newblock \emph{Advances in neural information processing systems}, 35:27730--27744.

\bibitem[{Pan et~al.(2024)Pan, Saxon, Xu, Nathani, Wang, and Wang}]{pan2024automatically}
Liangming Pan, Michael Saxon, Wenda Xu, Deepak Nathani, Xinyi Wang, and William~Yang Wang. 2024.
\newblock Automatically correcting large language models: Surveying the landscape of diverse automated correction strategies.
\newblock \emph{Transactions of the Association for Computational Linguistics}, 12:484--506.

\bibitem[{Patel et~al.(2021)Patel, Bhattamishra, and Goyal}]{patel-etal-2021-nlp}
Arkil Patel, Satwik Bhattamishra, and Navin Goyal. 2021.
\newblock \href {https://doi.org/10.18653/v1/2021.naacl-main.168} {Are {NLP} models really able to solve simple math word problems?}
\newblock In \emph{Proceedings of the 2021 Conference of the North American Chapter of the Association for Computational Linguistics: Human Language Technologies}, pages 2080--2094, Online. Association for Computational Linguistics.

\bibitem[{Paul et~al.(2024{\natexlab{a}})Paul, Ismayilzada, Peyrard, Borges, Bosselut, West, and Faltings}]{paul2024refiner}
Debjit Paul, Mete Ismayilzada, Maxime Peyrard, Beatriz Borges, Antoine Bosselut, Robert West, and Boi Faltings. 2024{\natexlab{a}}.
\newblock Refiner: Reasoning feedback on intermediate representations.
\newblock In \emph{Proceedings of the 18th Conference of the European Chapter of the Association for Computational Linguistics (Volume 1: Long Papers)}, pages 1100--1126.

\bibitem[{Paul et~al.(2024{\natexlab{b}})Paul, Ismayilzada, Peyrard, Borges, Bosselut, West, and Faltings}]{paul-etal-2024-refiner}
Debjit Paul, Mete Ismayilzada, Maxime Peyrard, Beatriz Borges, Antoine Bosselut, Robert West, and Boi Faltings. 2024{\natexlab{b}}.
\newblock \href {https://aclanthology.org/2024.eacl-long.67} {{REFINER}: Reasoning feedback on intermediate representations}.
\newblock In \emph{Proceedings of the 18th Conference of the European Chapter of the Association for Computational Linguistics (Volume 1: Long Papers)}, pages 1100--1126, St. Julian{'}s, Malta. Association for Computational Linguistics.

\bibitem[{Qiu et~al.(2024)Qiu, Jiang, Lu, Sclar, Pyatkin, Bhagavatula, Wang, Kim, Choi, Dziri et~al.}]{qiuphenomenal}
Linlu Qiu, Liwei Jiang, Ximing Lu, Melanie Sclar, Valentina Pyatkin, Chandra Bhagavatula, Bailin Wang, Yoon Kim, Yejin Choi, Nouha Dziri, et~al. 2024.
\newblock Phenomenal yet puzzling: Testing inductive reasoning capabilities of language models with hypothesis refinement.
\newblock In \emph{The Twelfth International Conference on Learning Representations}.

\bibitem[{Rae et~al.(2021)Rae, Borgeaud, Cai, Millican, Hoffmann, Song, Aslanides, Henderson, Ring, Young et~al.}]{rae2021scaling}
Jack~W Rae, Sebastian Borgeaud, Trevor Cai, Katie Millican, Jordan Hoffmann, Francis Song, John Aslanides, Sarah Henderson, Roman Ring, Susannah Young, et~al. 2021.
\newblock Scaling language models: Methods, analysis \& insights from training gopher.
\newblock \emph{arXiv preprint arXiv:2112.11446}.

\bibitem[{Shinn et~al.(2023)Shinn, Labash, and Gopinath}]{shinn2023reflexion}
Noah Shinn, Beck Labash, and Ashwin Gopinath. 2023.
\newblock Reflexion: an autonomous agent with dynamic memory and self-reflection.
\newblock \emph{arXiv preprint arXiv:2303.11366}, 2(5):9.

\bibitem[{Srivastava et~al.(2023)Srivastava, Rastogi, Rao, Shoeb, Abid, Fisch, Brown, Santoro, Gupta, Garriga-Alonso et~al.}]{srivastava2023beyond}
Aarohi Srivastava, Abhinav Rastogi, Abhishek Rao, Abu Awal~Md Shoeb, Abubakar Abid, Adam Fisch, Adam~R Brown, Adam Santoro, Aditya Gupta, Adri{\`a} Garriga-Alonso, et~al. 2023.
\newblock Beyond the imitation game: Quantifying and extrapolating the capabilities of language models.
\newblock \emph{Transactions on Machine Learning Research}.

\bibitem[{Stechly et~al.(2023)Stechly, Marquez, and Kambhampati}]{stechly2023gpt}
Kaya Stechly, Matthew Marquez, and Subbarao Kambhampati. 2023.
\newblock Gpt-4 doesn’t know it’s wrong: An analysis of iterative prompting for reasoning problems.
\newblock In \emph{NeurIPS 2023 Foundation Models for Decision Making Workshop}.

\bibitem[{Touvron et~al.(2023)Touvron, Lavril, Izacard, Martinet, Lachaux, Lacroix, Rozi{\`e}re, Goyal, Hambro, Azhar et~al.}]{touvron2023llama}
Hugo Touvron, Thibaut Lavril, Gautier Izacard, Xavier Martinet, Marie-Anne Lachaux, Timoth{\'e}e Lacroix, Baptiste Rozi{\`e}re, Naman Goyal, Eric Hambro, Faisal Azhar, et~al. 2023.
\newblock Llama: Open and efficient foundation language models.
\newblock \emph{arXiv preprint arXiv:2302.13971}.

\bibitem[{Valmeekam et~al.(2023)Valmeekam, Marquez, and Kambhampati}]{valmeekam2023can}
Karthik Valmeekam, Matthew Marquez, and Subbarao Kambhampati. 2023.
\newblock Can large language models really improve by self-critiquing their own plans?
\newblock In \emph{NeurIPS 2023 Foundation Models for Decision Making Workshop}.

\bibitem[{Wang et~al.(2023{\natexlab{a}})Wang, Xu, Lan, Hu, Lan, Lee, and Lim}]{wang-etal-2023-plan}
Lei Wang, Wanyu Xu, Yihuai Lan, Zhiqiang Hu, Yunshi Lan, Roy Ka-Wei Lee, and Ee-Peng Lim. 2023{\natexlab{a}}.
\newblock \href {https://doi.org/10.18653/v1/2023.acl-long.147} {Plan-and-solve prompting: Improving zero-shot chain-of-thought reasoning by large language models}.
\newblock In \emph{Proceedings of the 61st Annual Meeting of the Association for Computational Linguistics (Volume 1: Long Papers)}, pages 2609--2634, Toronto, Canada. Association for Computational Linguistics.

\bibitem[{Wang et~al.(2023{\natexlab{b}})Wang, Yu, Tan, O'Brien, Pasunuru, Dwivedi-Yu, Golovneva, Zettlemoyer, Fazel-Zarandi, and Celikyilmaz}]{wang2023shepherd}
Tianlu Wang, Ping Yu, Xiaoqing~Ellen Tan, Sean O'Brien, Ramakanth Pasunuru, Jane Dwivedi-Yu, Olga Golovneva, Luke Zettlemoyer, Maryam Fazel-Zarandi, and Asli Celikyilmaz. 2023{\natexlab{b}}.
\newblock Shepherd: A critic for language model generation.
\newblock \emph{arXiv preprint arXiv:2308.04592}.

\bibitem[{Wang et~al.(2023{\natexlab{c}})Wang, Wei, Schuurmans, Le, Chi, Narang, Chowdhery, and Zhou}]{wangself}
Xuezhi Wang, Jason Wei, Dale Schuurmans, Quoc~V Le, Ed~H Chi, Sharan Narang, Aakanksha Chowdhery, and Denny Zhou. 2023{\natexlab{c}}.
\newblock Self-consistency improves chain of thought reasoning in language models.
\newblock In \emph{The Eleventh International Conference on Learning Representations}.

\bibitem[{Wei et~al.(2022{\natexlab{a}})Wei, Tay, Bommasani, Raffel, Zoph, Borgeaud, Yogatama, Bosma, Zhou, Metzler et~al.}]{wei2022emergent}
Jason Wei, Yi~Tay, Rishi Bommasani, Colin Raffel, Barret Zoph, Sebastian Borgeaud, Dani Yogatama, Maarten Bosma, Denny Zhou, Donald Metzler, et~al. 2022{\natexlab{a}}.
\newblock Emergent abilities of large language models.
\newblock \emph{Transactions on Machine Learning Research}.

\bibitem[{Wei et~al.(2022{\natexlab{b}})Wei, Wang, Schuurmans, Bosma, Xia, Chi, Le, Zhou et~al.}]{wei2022chain}
Jason Wei, Xuezhi Wang, Dale Schuurmans, Maarten Bosma, Fei Xia, Ed~Chi, Quoc~V Le, Denny Zhou, et~al. 2022{\natexlab{b}}.
\newblock Chain-of-thought prompting elicits reasoning in large language models.
\newblock \emph{Advances in neural information processing systems}, 35:24824--24837.

\bibitem[{Welleck et~al.(2023)Welleck, Lu, West, Brahman, Shen, Khashabi, and Choi}]{welleck2023generating}
Sean Welleck, Ximing Lu, Peter West, Faeze Brahman, Tianxiao Shen, Daniel Khashabi, and Yejin Choi. 2023.
\newblock Generating sequences by learning to self-correct.
\newblock In \emph{The Eleventh International Conference on Learning Representations}.

\bibitem[{Xi et~al.(2023)Xi, Jin, Zhou, Zheng, Gao, Gui, Zhang, and Huang}]{xi2023self}
Zhiheng Xi, Senjie Jin, Yuhao Zhou, Rui Zheng, Songyang Gao, Tao Gui, Qi~Zhang, and Xuanjing Huang. 2023.
\newblock Self-polish: Enhance reasoning in large language models via problem refinement.
\newblock \emph{arXiv preprint arXiv:2305.14497}.

\bibitem[{Xiong et~al.(2024)Xiong, Ding, Liu, Qin, Xu, Yang, Liu, and Cao}]{xiong2024meaningful}
Kai Xiong, Xiao Ding, Ting Liu, Bing Qin, Dongliang Xu, Qing Yang, Hongtao Liu, and Yixin Cao. 2024.
\newblock Meaningful learning: Advancing abstract reasoning in large language models via generic fact guidance.
\newblock \emph{arXiv preprint arXiv:2403.09085}.

\bibitem[{Yang et~al.(2024)Yang, Yu, Zhang, Cao, Xu, Zhang, Gonzalez, and Cui}]{yang2024buffer}
Ling Yang, Zhaochen Yu, Tianjun Zhang, Shiyi Cao, Minkai Xu, Wentao Zhang, Joseph~E Gonzalez, and Bin Cui. 2024.
\newblock Buffer of thoughts: Thought-augmented reasoning with large language models.
\newblock \emph{arXiv preprint arXiv:2406.04271}.

\bibitem[{Yin et~al.(2023)Yin, Sun, Chang, Guo, Dai, Huang, and Qiu}]{yin-etal-2023-exchange}
Zhangyue Yin, Qiushi Sun, Cheng Chang, Qipeng Guo, Junqi Dai, Xuanjing Huang, and Xipeng Qiu. 2023.
\newblock \href {https://doi.org/10.18653/v1/2023.emnlp-main.936} {Exchange-of-thought: Enhancing large language model capabilities through cross-model communication}.
\newblock In \emph{Proceedings of the 2023 Conference on Empirical Methods in Natural Language Processing}, pages 15135--15153, Singapore. Association for Computational Linguistics.

\bibitem[{Zeng et~al.(2024)Zeng, Chen, Liu, Jiang, and Jia}]{zeng2024mrgsm8kmetareasoningbenchmarklarge}
Zhongshen Zeng, Pengguang Chen, Shu Liu, Haiyun Jiang, and Jiaya Jia. 2024.
\newblock \href {https://arxiv.org/abs/2312.17080} {Mr-gsm8k: A meta-reasoning benchmark for large language model evaluation}.
\newblock \emph{Preprint}, arXiv:2312.17080.

\bibitem[{Zhang et~al.(2024)Zhang, Shen, Wu, Peng, Wang, Zhuang, and Lu}]{zhang-etal-2024-self-contrast}
Wenqi Zhang, Yongliang Shen, Linjuan Wu, Qiuying Peng, Jun Wang, Yueting Zhuang, and Weiming Lu. 2024.
\newblock \href {https://doi.org/10.18653/v1/2024.acl-long.197} {Self-contrast: Better reflection through inconsistent solving perspectives}.
\newblock In \emph{Proceedings of the 62nd Annual Meeting of the Association for Computational Linguistics (Volume 1: Long Papers)}, pages 3602--3622, Bangkok, Thailand. Association for Computational Linguistics.

\bibitem[{Zheng et~al.(2024)Zheng, Mishra, Chen, Cheng, Chi, Le, and Zhou}]{zhengtake}
Huaixiu~Steven Zheng, Swaroop Mishra, Xinyun Chen, Heng-Tze Cheng, Ed~H Chi, Quoc~V Le, and Denny Zhou. 2024.
\newblock Take a step back: Evoking reasoning via abstraction in large language models.
\newblock In \emph{The Twelfth International Conference on Learning Representations}.

\end{thebibliography}

\appendix
\section{Datasets and Evaluation Metrics}
\label{sec:Datasets and Evaluation Metrics}

\textbf{Datasets} Table~\ref{table:dataset} presents detailed information about the datasets used in our experiments, including the data source, total number of test samples, task type, answer type, and the number of samples used to generate meta-thoughts, initial responses and reflection prompts.

\textbf{Evaluation Metrics}
Accuracy is used as the evaluation metric in our experiments. we use various methods to generate responses and exact answers in different tasks. For StrategyQA , we utilize Chain of Thought (CoT) to produce the response, then use regular expressions to extract the option following the phrase "so the answer is" and check if the extracted option matches the correct answer. For GSM8K and SVAMP, we employ the Program of Thought (PoT) to generate code for problem solving, which is then executed using a code executor to obtain the results. If the code contains errors, the program will raise an error and we define the results as None. Finally, we utilize the accuracy based on whether the results match the standard answers.
\section{Complementary Experiments Results}

\begin{table}[htbp]
\centering
\renewcommand{\arraystretch}{1.4}  
\small
\begin{adjustbox}{max width=\textwidth}  
\begin{tabular}{>{\centering\arraybackslash}p{1.0cm}|>{\centering\arraybackslash}p{1.0cm}|>{\centering\arraybackslash}p{0.7cm}>{\centering\arraybackslash}p{0.7cm}>{\centering\arraybackslash}p{0.7cm}>{\centering\arraybackslash}p{0.7cm}}  
\hline
{\centering GSM8K}&{\centering Iteration}&\textcolor{green}{\ding{51}}$\Rightarrow$\textcolor{green}{\ding{51}} & \textcolor{green}{\ding{51}}$\Rightarrow$\textbf{\textcolor{red}{\ding{55}}}&\textbf{\textcolor{red}{\ding{55}}}$\Rightarrow$\textcolor{green}{\ding{51}} & \textbf{\textcolor{red}{\ding{55}}}$\Rightarrow$\textbf{\textcolor{red}{\ding{55}}} \\
\hline
\multirow{4}{*}{\centering IoRT} 
& 0$\Rightarrow$1 & $76.6\%$ & $1.7\%$ & $5.9\%$ & $15.8\%$ \\
& 1$\Rightarrow$2 & $80.7\%$ & $1.8\%$ & $3.5\%$ & $14.0\%$ \\
& 2$\Rightarrow$3 & $82.7\%$ & $1.5\%$ & $1.7\%$ & $14.1\%$ \\
& 3$\Rightarrow$4 & $83.3\%$ & $1.1\%$ & $1.3\%$ & $14.3\%$ \\
\bottomrule
\end{tabular}
\end{adjustbox}
\caption{Performance of IoRT with GPT-3.5 during iterations on GSM8K.}
\label{table:IoRT}
\end{table}

\begin{table}[htbp]
\centering

\renewcommand{\arraystretch}{1.4}  
\small
\begin{adjustbox}{max width=\textwidth}  
\begin{tabular}{>{\centering\arraybackslash}p{1.4cm}|>{\centering\arraybackslash}p{1.0cm}|>{\centering\arraybackslash}p{0.7cm}>{\centering\arraybackslash}p{0.7cm}>{\centering\arraybackslash}p{0.7cm}>{\centering\arraybackslash}p{0.7cm}}  
\hline
{\centering GSM8K}&{\centering Iteration}&\textcolor{green}{\ding{51}}$\Rightarrow$\textcolor{green}{\ding{51}} & \textcolor{green}{\ding{51}}$\Rightarrow$\textbf{\textcolor{red}{\ding{55}}}&\textbf{\textcolor{red}{\ding{55}}}$\Rightarrow$\textcolor{green}{\ding{51}} & \textbf{\textcolor{red}{\ding{55}}}$\Rightarrow$\textbf{\textcolor{red}{\ding{55}}} \\
\hline
\multirow{4}{*}{\centering Self-Correct} 
& 0$\Rightarrow$1 & $71.2\%$ & $7.1\%$ & $5.1\%$ & $16.6\%$ \\
& 1$\Rightarrow$2 & $71.8\%$ & $4.5\%$ & $4.9\%$ & $18.8\%$ \\
& 2$\Rightarrow$3 & $72.3\%$ & $4.4\%$ & $3.0\%$ & $20.3\%$ \\
& 3$\Rightarrow$4 & $73.0\%$ & $2.3\%$ & $2.8\%$ & $21.9\%$ \\
\hline
\multirow{4}{*}{\centering CRITIC} 
& 0$\Rightarrow$1 & $71.0\%$ & $7.3\%$ & $4.9\%$ & $16.8\%$ \\
& 1$\Rightarrow$2 & $73.3\%$ & $2.6\%$ & $4.5\%$ & $19.6\%$ \\
& 2$\Rightarrow$3 & $74.3\%$ & $3.5\%$ & $2.5\%$ & $19.7\%$ \\
& 3$\Rightarrow$4 & $75.4\%$ & $1.4\%$ & $1.9\%$ & $21.3\%$ \\
\bottomrule
\end{tabular}
\end{adjustbox}
\caption{Performance of self-correct and CRITIC with GPT-3.5 during iterations on GSM8K.}
\label{table:critic} 
\end{table}

\begin{table*}[h]
\centering

\renewcommand{\arraystretch}{1.2}
\begin{tabular}{lcccccccl}
\hline
\textbf{Dataset}  & \textbf{Num.}& \textbf{Domain}&\textbf{Ans Type}&\textbf{$\#$ Meta}& \textbf{$\#$ Refresh} &\textbf{$\#$ Reflect} \\
\hline
GSM8K & 1,319  & Mathematical Reasoning&Number&8&8&4 \\
SVAMP & 1,000  & Mathematical Reasoning&Number&8&8&4\\ \hline
StrategyQA & 2,290  & Commonsense Reasoning&T/F&6&5&3 \\
\hline
\end{tabular}
\caption{Detailed statistics of the datasets used in our experiment. Num. indicates the number of samples in each dataset. Ans Type describes the form of the answer. $\#$Meta represents the number of few-shot examples used for generating meta-thoughts.  $\#$ Refresh refers to the number of chain-of-thought exemplars used as few-shot prompts for commonsense reasoning tasks, or program-of-thought exemplars for mathematical reasoning tasks. $\#$ Reflect represents the number of reflection exemplars used as few-shot prompts.}
\label{table:dataset}
\end{table*}

\section{Prompting Template}

\subsection{Prompt for Generate Meta Thoughts}

\begin{table}[H]
\centering
\begin{tikzpicture}
    \node [
        rounded corners=5pt, 
        draw=black, 
        line width=1.2pt, 
        fill=white, 
        inner sep=0pt, 
        outer sep=0pt, 
        minimum width=0.48\linewidth-2pt 
    ] (table) {
        \renewcommand{\arraystretch}{1.2} 
        \begin{tabular}{p{\dimexpr0.95\linewidth-4pt\relax}}
            \textbf{Generate Meta Thoughts} \\ 

You are a meta-thinker, skilled in abstract reasoning. Given a question, you should generate a meta-thought including the necessary knowledge, analytical methods, and fundamental strategies for solving the provided question. \\
             \textbf{Meta-thought Example:} [$\{e^{i}_{\text{sim}}\}$]\\
            \textbf{Question:} [Question x] \\
            \textbf{Meta-thought:} 
        \end{tabular}
    };
    
    \draw[line width=1.5pt] 
        (table.west |- 0,1.8) -- (table.east |- 0,1.8); 
\end{tikzpicture}
\label{tab:entities_relation_extraction1} 
\end{table}

\subsection{Prompt for Refresh and Self-Reflect}

\begin{table}[H]
\centering
\begin{tikzpicture}
    \node [
        rounded corners=5pt, 
        draw=black, 
        line width=1.2pt, 
        fill=white, 
        inner sep=0pt, 
        outer sep=0pt, 
        minimum width=0.48\linewidth-2pt 
    ] (table) {
        \renewcommand{\arraystretch}{1.4} 
        \begin{tabular}{p{\dimexpr0.95\linewidth-4pt\relax}}
            \textbf{Refresh (Mathematical Reasoning)} \\ 
            You are a mathematics and programming expert. Given a question, you should write python code to solve the following questions. Store your result as a variable named 'answer'.\\
            
            \textbf{Question:} [Question x] \\
            \textbf{Initial Response:} 
        \end{tabular}
    };
    
    \draw[line width=1.5pt] 
        (table.west |- 0,1.3) -- (table.east |- 0,1.3); 
\end{tikzpicture}
\label{tab:entities_relation_extraction2} 
\end{table}

\begin{table}[H]
\centering
\begin{tikzpicture}
    \node [
        rounded corners=5pt, 
        draw=black, 
        line width=1.2pt, 
        fill=white, 
        inner sep=0pt, 
        outer sep=0pt, 
        minimum width=0.48\linewidth-2pt 
    ] (table) {
        \renewcommand{\arraystretch}{1.4} 
        \begin{tabular}{p{\dimexpr0.95\linewidth-4pt\relax}}
            \textbf{Self-Reflect  (Mathematical Reasoning)} \\ 
            You are a mathematics and programming expert. Given previous responses, including a question, a python code solution and the output of this code, you should carefully evaluate these responses and provide detailed feedback. \\
            
            \textbf{Question:} [Question x]\\
            \textbf{Basic Response:} [$R^{i}_{b}$]\\
            \textbf{Basic Answer:} [$A^{i}_{b}$]\\
            \textbf{Evaluation Feedback:} \\
            You are a mathematics and programming expert. Based on evaluation feedback including a question, a python code solution, the output of this code, and the code evaluation, you should generate a better code based on the feedback.\\
            
            \textbf{Question:} [Question x]\\
            \textbf{Basic Response:} [$R^{i}_{b}$]\\
            \textbf{Basic Answer:} [$A^{i}_{b}$]\\
            \textbf{Evaluation Feedback:} [$f_i$]\\
            \textbf{Reflect Response:}  \\
        \end{tabular}
    };
    
    \draw[line width=1.5pt] 
        (table.west |- 0,5.2) -- (table.east |- 0,5.2); 
    \draw[line width=1.5pt] 
        (table.west |- 0,0) -- (table.east |- 0,0); 
\end{tikzpicture}
\label{tab:entities_relation_extraction3} 
\end{table}

\begin{table}[H]
\centering
\begin{tikzpicture}
    \node [
        rounded corners=5pt, 
        draw=black, 
        line width=1.2pt, 
        fill=white, 
        inner sep=0pt, 
        outer sep=0pt, 
        minimum width=0.48\linewidth-2pt 
    ] (table) {
        \renewcommand{\arraystretch}{1.4} 
        \begin{tabular}{p{\dimexpr0.95\linewidth-4pt\relax}}
            \textbf{Refresh (Commonsense Reasoning)} \\ 
            You are a knowledgeable expert in general knowledge and common sense. Given you a question, you should think step by step and then generate the answer. \\
            
            \textbf{Question:} [Question x] \\
            \textbf{Initial Response:} 
        \end{tabular}
    };
    
    \draw[line width=1.5pt] 
        (table.west |- 0,1.3) -- (table.east |- 0,1.3); 
\end{tikzpicture}
\label{tab:entities_relation_extraction4} 
\end{table}

\begin{table}[H]
\centering
\begin{tikzpicture}
    \node [
        rounded corners=5pt, 
        draw=black, 
        line width=1.2pt, 
        fill=white, 
        inner sep=0pt, 
        outer sep=0pt, 
        minimum width=0.48\linewidth-2pt 
    ] (table) {
        \renewcommand{\arraystretch}{1.5} 
        \begin{tabular}{p{\dimexpr0.95\linewidth-4pt\relax}}
            \textbf{Self-Reflect  (Commonsense Reasoning)} \\ 
            You are a general knowledge and common sense expert. Given previous responses, which include a question, a reasoning chain and the answer, you should carefully evaluate these responses and provide detailed feedback. \\
            
            \textbf{Question:} [Question x]\\
            \textbf{Basic Response:} [$R^{i}_{b}$]\\
            \textbf{Basic Answer:} [$A^{i}_{b}$]\\
            \textbf{Evaluation Feedback:} \\
    
            You are a general knowledge and common sense expert. Based on evaluation feedback including a question, a reasoning chain, a answer and the evaluation feedback, you should generate a better reasoning chain  based on the feedback.\\
            \textbf{Question:} [Question x]\\
            \textbf{Basic Response:} [$R^{i}_{b}$]\\
            \textbf{Basic Answer:} [$A^{i}_{b}$]\\
            \textbf{Evaluation Feedback:} [$f_i$]\\
            \textbf{Reflect Response:} \\
        \end{tabular}
    };
    
    \draw[line width=1.5pt] 
        (table.west |- 0,5.75) -- (table.east |- 0,5.75); 
    \draw[line width=1.5pt] 
        (table.west |- 0,0.25) -- (table.east |- 0,0.25); 
\end{tikzpicture}
\label{tab:entities_relation_extraction5} 
\end{table}

\subsection{Prompt for Instruct the Iteration}

\begin{table}[H]
\centering
\begin{tikzpicture}
    \node [
        rounded corners=5pt, 
        draw=black, 
        line width=1.2pt, 
        fill=white, 
        inner sep=0pt, 
        outer sep=0pt, 
        minimum width=0.48\linewidth-2pt 
    ] (table) {
        \renewcommand{\arraystretch}{1.8} 
        \begin{tabular}{p{\dimexpr0.85\linewidth-4pt\relax}}
            \textbf{Select instruction}\\  \textbf{(Mathematical Reasoning)} \\ 
             The outputs from the two codes are different. Evaluate the following two Python code snippets based on the given question, output of the code and meta-thought, and recommend which code is better to solve this question and explain why. \\
            \textbf{Question:} [Question x]\\
            \textbf{Meta Thought:} [$m_i$]\\
            \textbf{Basic Response:} [$R^{i}_{b}$]\\
            \textbf{Basic Answer:} [$A^{i}_{b}$]\\
            \textbf{Reflect Response:} [$R^{i}_{r}$]\\
             \textbf{Reflect Answer:} [$A^{i}_{r}$]\\
            \textbf{Better Response:} \\
        \end{tabular}
    };
    
    \draw[line width=1.5pt] 
        (table.west |- 0,3.8) -- (table.east |- 0,3.8); 
   
\end{tikzpicture}
\label{tab:entities_relation_extraction6} 
\end{table}

\begin{table}[H]
\centering
\begin{tikzpicture}
    \node [
        rounded corners=5pt, 
        draw=black, 
        line width=1.2pt, 
        fill=white, 
        inner sep=0pt, 
        outer sep=0pt, 
        minimum width=0.48\linewidth-2pt 
    ] (table) {
        \renewcommand{\arraystretch}{1.8} 
        \begin{tabular}{p{\dimexpr0.85\linewidth-4pt\relax}}
            \textbf{Select instruction}\\  \textbf{(Commonsense Reasoning)} \\ 
            Evaluate the following two reasoning chain based on the given question and meta though. Recommend which reasoning chain is better and explain why. \\
            \textbf{Question:} [Question x]\\
            \textbf{Meta Thought:} [$m_i$]\\
            \textbf{Basic Response:} [$R^{i}_{b}$]\\
            \textbf{Basic Answer:} [$A^{i}_{b}$]\\
            \textbf{Reflect Response:} [$R^{i}_{r}$]\\
            \textbf{Reflect Answer:} [$A^{i}_{r}$]\\
            \textbf{Better Response:} \\
            
        \end{tabular}
    };
    
    \draw[line width=1.5pt] 
        (table.west |- 0,3.2) -- (table.east |- 0,3.2); 
   
\end{tikzpicture}
\label{tab:entities_relation_extraction7} 
\end{table}

\begin{table}[H]
\centering
\begin{tikzpicture}
    \node [
        rounded corners=5pt, 
        draw=black, 
        line width=1.2pt, 
        fill=white, 
        inner sep=0pt, 
        outer sep=0pt, 
        minimum width=0.48\linewidth-2pt 
    ] (table) {
        \renewcommand{\arraystretch}{1.8} 
        \begin{tabular}{p{\dimexpr0.85\linewidth-4pt\relax}}
            \textbf{Refresh or Stop  instruction} \\ \textbf{(Mathematical Reasoning)} \\ 
            The outputs from the two codes are the same. Please evaluate the following two Python code snippets based on the question, output of the code and meta thought. Decide whether to refresh a new solution or to stop. \\
            \textbf{Question:} [Question x]\\
            \textbf{Meta Thought:} [$m_i$]\\
            \textbf{Basic Response:} [$R^{i}_{b}$]\\
             \textbf{Basic Answer:} [$A^{i}_{b}$]\\
            \textbf{Reflect Response:} [$R^{i}_{r}$]\\
             \textbf{Reflect Answer:} [$A^{i}_{r}$]\\
            \textbf{Decision:} \\
        \end{tabular}
    };
    
    \draw[line width=1.5pt] 
        (table.west |- 0,3.7) -- (table.east |- 0,3.7); 
   
\end{tikzpicture}
\label{tab:entities_relation_extraction8} 
\end{table}

\begin{table}[H]
\centering
\begin{tikzpicture}
    \node [
        rounded corners=5pt, 
        draw=black, 
        line width=1.2pt, 
        fill=white, 
        inner sep=0pt, 
        outer sep=0pt, 
        minimum width=0.48\linewidth-2pt 
    ] (table) {
        \renewcommand{\arraystretch}{1.8} 
        \begin{tabular}{p{\dimexpr0.85\linewidth-4pt\relax}}
            \textbf{Refresh or Stop  instruction} \\ \textbf{(Commonsense Reasoning)} \\ 
            The answers from the reasoning chains are the same. Please evaluate these two reasoning chains based on the following question and meta thought. Decide whether to refresh a new solution or to stop. \\
            \textbf{Question:} [Question x]\\
            \textbf{Meta Thought:} [$m_i$]\\
            \textbf{Basic Response:} [$R^{i}_{b}$]\\
             \textbf{Basic Answer:} [$A^{i}_{b}$]\\
            \textbf{Reflect Response:} [$R^{i}_{r}$]\\
             \textbf{Reflect Answer:} [$A^{i}_{r}$]\\
            \textbf{Better Response:} \\
        \end{tabular}
    };
    
    \draw[line width=1.5pt] 
        (table.west |- 0,3.5) -- (table.east |- 0,3.5); 
   
\end{tikzpicture}
\label{tab:entities_relation_extraction9} 
\end{table}
\onecolumn
\section{Case Study}
To better understand the IoRT working mechanism, we analyze cases in different scenarios.

\begin{table*}[h]
\centering
\renewcommand{\arraystretch}{1.3}
\begin{tabular}{p{0.85\textwidth}|p{0.1\textwidth}}
\hline
\multicolumn{1}{c|}{\textbf{Iterative Reflection Process}} & \multicolumn{1}{c}{\textbf{Output}} \\
\hline
\textbf{Question:} Would a veteran of the Phillippine-American War come home craving SPAM? & \multirow{9}{=}{\centering Iteration 0\\False (\textcolor{green}{\ding{51}})} \\
\textbf{Generate Meta-thoughts:} & \\
Research when SPAM was introduced to the market and compare that to the timeline of the Philippine-American War to assess the likelihood of a veteran having access to or knowledge of SPAM upon returning home. & \\
\textbf{Initial Response:} & \\
Let's think step by step. The Philippine-American War occurred between 1899 and 1902, and SPAM was introduced in 1937. There was no direct relationship between these two events, so it's unlikely that a veteran of the Philippine-American War would come home craving SPAM. So the answer is: false.&\\
\textbf{Extract the Answer:} False (\textcolor{green}{\ding{51}}) CoT 0&\\
\hline
\textbf{Self-Reflect:} & \multirow{20}{=}{\centering Iteration 1 False (\textcolor{green}{\ding{51}})}\\

What's the problem with the above answer?&\\

1. Plausibility: The answer logically explains the lack of a direct relationship between the Philippine-American War and the introduction of SPAM in 1937, which makes it unlikely for a veteran of the war to come home craving SPAM. The reasoning is sound and aligns with historical timelines.&\\

2. Truthfulness: The information provided is accurate. The timeline of the events mentioned is correct, and there is indeed no direct connection between the Philippine-American War and the introduction of SPAM in 1937.&\\

Here's the better answer: Let's think step by step. The Philippine-American War took place between 1899 and 1902, while SPAM was introduced in 1937, which means there is no direct correlation between the two events. Therefore, it is unlikely that a veteran of the Philippine-American War would come home craving SPAM. So the answer is: false.&\\
\textbf{Extract the Answer:} False (\textcolor{green}{\ding{51}}) CoT 1&\\

\textbf{Instruct the Iteration (Stop):}&\\
Both COTs provided accurate and logical reasoning based on the question and meta thought provided. They correctly assessed the timeline of the events and concluded that there was no direct correlation between the Philippine-American War and the introduction of SPAM, making it unlikely for a veteran of that war to come home craving SPAM. Therefore, there is no need to regenerate a new answer as the current responses are appropriate.
Decision: Stop iteration.&\\
\hline

\end{tabular}
\caption{An illustrative example from StrategyQA demonstrating how IoRT evaluate the quality of responses and generates a stop instruction when the results before and after reflection are consistent.}
\label{table:case_study1}
\end{table*}

\begin{table*}[h]
\centering

\renewcommand{\arraystretch}{1.3}
\begin{tabular}{p{0.85\textwidth}|p{0.1\textwidth}}
\hline
\multicolumn{1}{c|}{\textbf{Iterative Reflection Process}} & \multicolumn{1}{c}{\textbf{Output}} \\
\hline
\textbf{Question:} Can you buy a fair trade laptop?
 & \multirow{9}{=}{\centering Iteration 0 \\True (\textbf{\textcolor{red}{\ding{55}}})} \\
\textbf{Generate Meta-thoughts:} & \\
Research the availability of fair trade electronics and the criteria for a product to be certified as fair trade. Consider the supply chains involved in laptop production and whether any companies offer laptops that meet fair trade standards. & \\
\textbf{Initial Response:} & \\
Let's think step by step. Fair trade principles are typically associated with agricultural or manufactured goods where workers are paid fair wages and operate in safe conditions. While there aren't fair trade-certified laptops per se, some companies may adhere to fair trade practices in the production of their electronics to ensure fair pay and safe working conditions for workers involved in their supply chain. However, it's less common in the electronics industry compared to agricultural products. So the answer is: true. &\\ 
\textbf{Extract the Answer:} True
(\textbf{\textcolor{red}{\ding{55}}}) CoT 0 & \\
\hline
\textbf{Self-Reflect:} & \multirow{23}{=}{\centering Iteration 1\\False (\textcolor{green}{\ding{51}})}\\

What's the problem with the above answer?&\\

1. Plausibility: & \\
The answer provides a reasonable explanation of fair trade principles and their application in the electronics industry. However, it does not definitively state whether fair trade laptops are available for purchase, leaving room for interpretation.& \\

2. Truthfulness: & \\
The information provided is accurate in explaining fair trade practices and their common association with agricultural products. While fair trade-certified laptops may not be prevalent, some companies do indeed prioritize fair trade principles in their electronics production to ensure ethical practices.& \\

Here's the better answer: &\\
Fair trade-certified laptops are not commonly available in the market due to the focus of fair trade principles on agricultural and manufactured goods. While some electronics companies may adhere to fair trade practices in their supply chain, specifically for components like minerals or labor conditions, the concept of fair trade laptops as a distinct certification is not widespread. So the answer is: false.&\\
\textbf{Extract the Answer:} False (\textcolor{green}{\ding{51}}) CoT 1&\\
&\\
\textbf{Instruct the Iteration (Select):}&\\
COT 1 provides a more accurate and precise response by directly addressing the question of fair trade-certified laptops and their availability in the market. It aligns closely with the meta thought provided and offers a clear 'false' conclusion based on the information presented. Better COT after comparison: COT 1 &\\
\hline

\end{tabular}
\caption{An illustrative example of the better response selection of IoRT in StrategyQA. When the results before and after reflection are inconsistent, a select instruction is triggered automatically and the instructor then chooses the better response.}
\label{table:case_study2}
\end{table*}

\begin{table*}[h]
\centering

\renewcommand{\arraystretch}{1.3}
\begin{tabular}{p{0.85\textwidth}|p{0.1\textwidth}}
\hline
\multicolumn{1}{c|}{\textbf{Iterative Reflection Process}} & \multicolumn{1}{c}{\textbf{Output}} \\
\hline
\textbf{Question:} Can a honey bee sting a human more than once? & \multirow{9}{=}{\centering Iteration 0 \\False (\textcolor{green}{\ding{51}})} \\
\textbf{Generate Meta-thoughts:} & \\
Research the physical structure of a honey bee's stinger and how it behaves when the bee stings a mammalian target. Consider what happens to the bee and its stinger after the stinging event. & \\
\textbf{Initial Response:} & \\
Let's think step by step. When a honey bee stings a human, its stinger gets embedded in the skin and is ripped off from the bee's body, causing the bee to die shortly after. So, honey bees can sting a human only once as they cannot survive after stinging. So the answer is: false.&\\
\textbf{Extract the Answer:} False (\textcolor{green}{\ding{51}}) CoT 0 & \\
\hline
\textbf{Self-Reflect:} & \multirow{23}{=}{\centering Iteration 1\\False (\textcolor{green}{\ding{51}})}\\

What's the problem with the above answer?&\\
1. Plausibility: &\\
The answer explains the mechanism of a honey bee's stinger getting embedded and the bee dying after stinging. While this information is correct, it does not fully address the question of whether a honey bee can sting a human more than once.&\\ 
2. Truthfulness: &\\
The information provided about a honey bee's stinger getting ripped off and the bee dying after stinging is accurate. However, the answer fails to consider that not all bees behave the same way. Honey bees are unique in this aspect, but other types of bees, such as bumblebees, can sting multiple times without dying.&\\
&\\
Here's the better answer: &\\
Honey bees can sting a human only once as their stinger gets embedded in the skin and is ripped off, causing the bee to die. However, it's important to note that not all bees behave this way. For example, bumblebees can sting multiple times without dying. So the answer is: true. &\\
\textbf{Extract the Answer:} True (\textbf{\textcolor{red}{\ding{55}}}) CoT 1&\\
&\\
\textbf{Instruct the Iteration (Select):}&\\
Based on the question and meta-thought provided, CoT 0 is recommended as the better choice because it provides a logical, and directly relevant answer to the question without introducing unnecessary information about other bee species. It adheres closely to the meta thought provided and offers a concise response based on the specific behavior of honey bees. Better COT after comparison: CoT 0 &\\
\hline

\end{tabular}
\caption{An illustrative example of the better response selection of IoRT in StrategyQA. When the results before and after reflection are inconsistent, a select instruction is triggered automatically and the instructor then chooses the better response.}
\label{table:case_study3}
\end{table*}

\begin{table*}[h]
\centering

\renewcommand{\arraystretch}{0.9}
\begin{tabular}{p{0.8\textwidth}|p{0.1\textwidth}}
\hline
\multicolumn{1}{c|}{\textbf{Iterative Reflection Process}} & \multicolumn{1}{c}{\textbf{Output}} \\
\hline
\textbf{Question:} Is a doctorate required to teach at a SUNY School? & \multirow{9}{=}{\centering Iteration 0 \\True (\textbf{\textcolor{red}{\ding{55}}})} \\
\textbf{Generate Meta-thoughts:} & \\
Research the general requirements for faculty positions at State University of New York (SUNY) schools and the typical educational qualifications expected for teaching at the college level, particularly in the subject area of interest. & \\
\textbf{Initial Response:} & \\
Let's think step by step. SUNY schools are part of the State University of New York system. Each SUNY school may have different requirements for their faculty positions, but generally, a doctorate is a common requirement for teaching at the university level. So the answer is: true. &\\
\textbf{Extract the Answer:} True (\textbf{\textcolor{red}{\ding{55}}}) CoT 0&\\
\hline
\textbf{Self-Reflect:} & \multirow{23}{=}{\centering Iteration 1\\True (\textbf{\textcolor{red}{\ding{55}}})}\\

What's the problem with the above answer?& \\

1. Plausibility: & \\
The answer logically addresses the question by explaining that while specific SUNY schools may have varying requirements, having a doctorate is generally a common requirement for teaching at the university level. This explanation aligns with the common standards in academia.& \\

2. Truthfulness: & \\
The information provided is accurate. While there may be exceptions based on specific disciplines or positions within SUNY schools, having a doctorate is indeed a common requirement for teaching at the university level, including SUNY schools.& \\

Here's the better answer: Generally, having a doctorate is a common requirement for teaching at the university level, including SUNY schools. So the answer is: true.&\\
\textbf{Extract the Answer:} True
(\textbf{\textcolor{red}{\ding{55}}}) CoT 1&\\
&\\
\textbf{Instruct the Iteration (Refresh):}&\\
Both COTs provide similar answers stating that a doctorate is generally required to teach at a SUNY school. However, the answers lack depth in addressing the specific requirements for faculty positions at SUNY schools and the educational qualifications expected for teaching at the college level. To improve the response, a more detailed analysis based on the meta thought provided is necessary.
Decision: Refreshing the response.&\\
\hline
\textbf{Refresh}&\multirow{12}{=}{\centering Iteration 2\\False (\textcolor{green}{\ding{51}})}\\
A doctorate is not always required to teach at a SUNY (State University of New York) school. While many professors at SUNY schools hold doctorate degrees, some instructors may have a master's degree or other qualifications depending on the specific requirements of the position and the subject being taught. So the answer is: false. &\\
\textbf{Extract the Answer:} False (\textcolor{green}{\ding{51}}) CoT 2&\\
&\\
\textbf{Instruct the Iteration (Select)}:& \\
Based on the question and meta-thought provided, COT 2 provides a more nuanced and accurate answer by considering the variability in educational qualifications for teaching at SUNY schools, which aligns better with the research and understanding of the requirements for faculty positions at SUNY schools.Better COT after comparison: COT 2\\
\hline
\end{tabular}
\caption{An illustrative example from StrategyQA demonstrating how IoRT generates a refresh instruction when the results before and after reflection are consistent.}
\label{table:case_study4}
\end{table*}

\begin{table*}[h]
\centering

\renewcommand{\arraystretch}{1.1}
\begin{tabular}{p{0.8\textwidth}|p{0.1\textwidth}}
\hline
\multicolumn{1}{c|}{\textbf{Iterative Reflection Process}} & \multicolumn{1}{c}{\textbf{Output}} \\
\hline
\textbf{Question:} A candle melts by 2 centimeters every hour that it burns. How many centimeters shorter will a candle be after burning from 1:00 PM to 5:00 PM? & \multirow{12}{=}{\centering Iteration 0\\8.0 (\textcolor{green}{\ding{51}})} \\
\textbf{Generate Meta-thoughts:} & \\
To determine how many centimeters shorter the candle will be after burning for a certain number of hours, you need to calculate the total time the candle burns and then multiply that by the rate at which the candle melts per hour. &\\

\textbf{Initial Response:} &\\

\text{burning\_hours} = 5 - 1 &\\ 
\text{melting\_rate\_per\_hour} = 2 &\\ 
answer = \text{burning\_hours} * \text{melting\_rate\_per\_hour} &\\ 
\textbf{Extract the Answer}: 8.0 
 (\textcolor{green}{\ding{51}}) Code 0&\\
\hline
\textbf{Self-Reflect:} & \multirow{20}{=}{\centering Iteration 1 8.0 (\textcolor{green}{\ding{51}})}\\

What's the problem with the above code? &\\

1. The output of 8.0 seems reasonable, as the candle would be 8 centimeters shorter after burning from 1:00 PM to 5:00 PM. &\\

2. Let's check the code: &\\

\text{burning\_hours = 5 - 1} &\\
\text{melting\_rate\_per\_hour = 2} &\\ 
\text{answer = burning\_hours * melting\_rate\_per\_hour} &\\

The code calculates the total burning hours and the melting rate per hour correctly, and then calculates the total length the candle would be shorter by multiplying the burning hours with the melting rate per hour. &\\

Overall, the code correctly calculates the length the candle would be shorter after burning from 1:00 PM to 5:00 PM. &\\
\\
\textbf{Here's a better code:} \\

\text{burning\_hours = 4} \\ 
\text{melting\_rate\_per\_hour = 2} \\ 
\text{answer = burning\_hours * melting\_rate\_per\_hour} &\\
\textbf{Extract the Answer: }8.0 (\textcolor{green}{\ding{51}}) Code 1
&\\
\textbf{Instruct the Iteration (Stop):}&\\
Both code snippets correctly calculate the number of centimeters the candle will be shorter after burning for the specified hours. The calculations are consistent with the hint provided, and the outputs match the expected result. There is no need to regenerate a new answer as the current solution is clear, logical, and directly addresses the question. Decision: stop iteration.
&\\
\hline
\end{tabular}
\caption{An illustrative example from GSM8K demonstrating how IoRT generates a stop instruction when the results before and after reflection are consistent.}
\label{table:case_study5}
\end{table*}

\begin{table*}[h]

\renewcommand{\arraystretch}{1.1}
\begin{tabular}{p{0.8\textwidth}|p{0.1\textwidth}}
\hline
\multicolumn{1}{c|}{\textbf{Iterative Reflection Process}} & \multicolumn{1}{c}{\textbf{Output}} \\
\hline
\textbf{Question:} Gloria is shoe shopping when she comes across a pair of boots that fit her shoe budget. However, she has to choose between the boots and two pairs of high heels that together cost five dollars less than the boots. If one pair of heels costs $\$$33 and the other costs twice as much, how many dollars are the boots?

 & \multirow{12}{=}{\centering Iteration 0 \\94.0 (\textbf{\textcolor{red}{\ding{55}}})} \\
\textbf{Generate Meta-thoughts:} & \\
To determine the cost of the boots, you need to calculate the total cost of the two pairs of high heels and then add five dollars to that amount. & \\
\textbf{Initial Response:} & \\
\text{cost\_of\_high\_heels\_1} = 33 \\ 
\text{cost\_of\_high\_heels\_2} = 2 * \text{cost\_of\_high\_heels\_1} \\ 
\text{total\_cost\_of\_high\_heels} = \text{cost\_of\_high\_heels\_1} + \text{cost\_of\_high\_heels\_2} \\ 
\text{difference} = 5 \\ 
\text{cost\_of\_boots} = \text{total\_cost\_of\_high\_heels} - \text{difference} \\ 
\text{answer} = \text{cost\_of\_boots} \\ 

\textbf{Extract the Answer:}  94.0 (\textbf{\textcolor{red}{\ding{55}}}) Code 0 & \\
\hline
\textbf{Self-Reflect:} & \multirow{23}{=}{\centering Iteration 1\\104.0 (\textcolor{green}{\ding{51}})}\\

What's the problem with the above answer?&\\

1. \text{The cost of the boots should be a positive number, } 94 > 0, \text{ it's reasonable.} \\

2. \text{Let's check the code:} \\

\text{cost\_of\_high\_heels\_1} = 33 \\ 
\text{cost\_of\_high\_heels\_2} = 2 * \text{cost\_of\_high\_heels\_1} \\ 
\text{total\_cost\_of\_high\_heels} = \text{cost\_of\_high\_heels\_1} + \text{cost\_of\_high\_heels\_2} \\

It calculates the total cost of the two pairs of high heels, that's correct. \\

\text{difference} = 5 \\ 
\text{cost\_of\_boots} = \text{total\_cost\_of\_high\_heels} - \text{difference} \\

\text{This calculates the cost of the boots, it's correct.} \\

In summary, the code correctly calculates the cost of the boots. \\

The code is correct and provides the expected output. \\
\\
\textbf{Here's a better code:} \\

\text{cost\_of\_high\_heels\_1} = 33 \\ 
\text{cost\_of\_high\_heels\_2} = 2 * \text{cost\_of\_high\_heels\_1} \\ 
\text{total\_cost\_of\_high\_heels} = \text{cost\_of\_high\_heels\_1} + \text{cost\_of\_high\_heels\_2} \\ 
\text{cost\_of\_boots} = \text{total\_cost\_of\_high\_heels} + 5 \\ 
\text{answer} = \text{cost\_of\_boots} \\

\text{Extract the Answer:} 104.0
 (\textcolor{green}{\ding{51}}) Code 1&\\
&\\
\textbf{Instruct the Iteration (Select):}&\\
Code 1 is more aligned with the hint provided as it correctly adds 5 dollars to the total cost of the two pairs of high heels to calculate the cost of the boots. This approach is consistent with the problem description and the meta thought given. Better Code after comparison: Code 1 &\\
\hline

\end{tabular}
\caption{An illustrative example of the better response selection of IoRT in GSM8K. When the results before and after reflection are inconsistent, a select instruction is triggered automatically and the instructor then chooses the better response.}
\label{table:case_study6}
\end{table*}

\begin{table*}[h]
\centering

\renewcommand{\arraystretch}{0.9}
\begin{tabular}{p{0.85\textwidth}|p{0.1\textwidth}}
\hline
\multicolumn{1}{c|}{\textbf{Iterative Reflection Process}} & \multicolumn{1}{c}{\textbf{Output}} \\
\hline
\textbf{Question:} Richard lives in an apartment building with 15 floors. Each floor contains 8 units, and 3/4 of the building is occupied. What's the total number of unoccupied units In the building? & \multirow{9}{=}{\centering Iteration 0 \\30.0 (\textcolor{green}{\ding{51}})} \\
\textbf{Generate Meta-thoughts:} & \\
To calculate the total number of unoccupied units in the building, first determine the total number of units in the building. Then, find the number of occupied units. Finally, calculate the number of unoccupied units. & \\
\textbf{Initial Response:} & \\
\text{total\_floors = 15} \\
\text{units\_per\_floor = 8} \\
\text{occupancy\_rate = 3/4} \\
\text{total\_units = total\_floors * units\_per\_floor} \\
\text{occupied\_units = total\_units * occupancy\_rate} \\
\text{unoccupied\_units = total\_units - occupied\_units} \\
\text{answer = unoccupied\_units} \\
\textbf{Extract the Answer:} 30.0
(\textcolor{green}{\ding{51}}) Code 0 & \\
\hline
\textbf{Self-Reflect:} & \multirow{23}{=}{\centering Iteration 1\\30.0 (\textcolor{green}{\ding{51}})}\\

What's the problem with the above answer?&\\
1. The total number of unoccupied units should be a whole number, 30.0 is not a whole number, so the answer is not reasonable.&\\
2. Let's check the code:&\\
\text{total\_floors = 15} \\
\text{units\_per\_floor = 8} \\
\text{occupancy\_rate = 3/4}\\

It calculates the total number of units, the number of occupied units, and then the number of unoccupied units, which is correct.\\
\text{answer = unoccupied\_units}\\
The code assigns the number of unoccupied units to the variable answer, which is correct.\\

The code seems to be correct, but the output is not a whole number. This could be due to the way the division is handled in Python. When dividing integers, Python returns a float if the result is not a whole number. To ensure we get a whole number as the output, we can use integer division.\\

\textbf{Here's a better code:}\\
\text{total\_floors = 15} \\
\text{units\_per\_floor = 8} \\
\text{occupancy\_rate = 3/4} \\
\text{total\_units = total\_floors * units\_per\_floor} \\
\text{occupied\_units = total\_units * occupancy\_rate} \\
\text{unoccupied\_units = total\_units - occupied\_units} \\
\text{answer = unoccupied\_units}\\

\textbf{Extract the Answer:} 120.0 (\textbf{\textcolor{red}{\ding{55}}}) Code 1&\\
&\\
\textbf{Instruct the Iteration (Select):}&\\
Code 0 is the recommended choice because it follows the hint accurately by using floating-point division to calculate the occupancy rate. The output of Code 0 aligns with the problem statement and the correct calculation methodology provided in the hint. On the other hand, Code 1 deviates from the correct calculation by using integer division, leading to an incorrect result for unoccupied units. Better Code after comparison: Code 0 &\\
\hline

\end{tabular}
\caption{An illustrative example of the better response selection of IoRT in GSM8K. When the results before and after reflection are inconsistent, a select instruction is triggered automatically and the instructor then chooses the better response.}
\label{table:case_study7}
\end{table*}

\onecolumn
\renewcommand{\arraystretch}{1.1}
\begin{longtable}{p{0.85\textwidth}|p{0.1\textwidth}}
   
    \hline
    \multicolumn{1}{c|}{\textbf{Iterative Reflection Process}} & \multicolumn{1}{c}{\textbf{Output}} \\

\hline
\textbf{Question:} Vincent can buy flowers in packages of 3 for $\$$2.50 or in packages of 2 for $\$$1. How much money does he save by buying 18 flowers at the better price?
 & \multirow{9}{=}{\centering Iteration 0 \\3.0 (\textbf{\textcolor{red}{\ding{55}}})} \\
\textbf{Generate Meta-thoughts:} & \\
To determine the savings Vincent can achieve by buying flowers at the better price, you need to calculate the cost per flower for each package option and then multiply the cost of the cheaper option by the total number of flowers to find the total cost. & \\
\textbf{Initial Response:} & \\
\text{num\_flowers} = 18 & \\
\text{price\_3\_flowers} = 2.5 & \\
\text{price\_2\_flowers} = 1 & \\
\text{total\_price\_3\_flowers} = (\text{num\_flowers} // 3)*\text{price\_3\_flowers} & \\
\text{remaining\_flowers} = \text{num\_flowers} \% 3 & \\
\text{total\_price\_2\_flowers} = (\text{remaining\_flowers} // 2) * \text{price\_2\_flowers} & \\
\text{answer}=(\text{num\_flowers}*\text{price\_2\_flowers})-(\text{total\_price\_3\_flowers}+\text{total\_price\_2\_flowers}) & \\
\textbf{Extract the Answer:} 3.0 
(\textbf{\textcolor{red}{\ding{55}}}) Code 0&\\
\hline
\textbf{Self-Reflect:} & \multirow{23}{=}{\centering Iteration 1\\3.0 (\textbf{\textcolor{red}{\ding{55}}})}\\

What's the problem with the above answer?& \\

1. The amount of money saved should be a positive number, \text{3 > 0}, it's reasonable.

2. Let's check the code:

> \text{num\_flowers} = 18
> \text{price\_3\_flowers} = 2.5
> \text{price\_2\_flowers} = 1

These lines define the number of flowers and the prices for different packages, that's correct.

> \text{total\_price\_3\_flowers} = (\text{num\_flowers} // 3) * \text{price\_3\_flowers}

This line calculates the total cost of buying flowers in packages of 3, it's correct.

> \text{remaining\_flowers} = \text{num\_flowers} \% 3
> \text{total\_price\_2\_flowers} = (\text{remaining\_flowers} // 2) * \text{price\_2\_flowers}

These lines calculate the total cost of buying flowers in packages of 2 for the remaining flowers, it's correct.

> \text{answer} = (\text{num\_flowers}*\text{price\_2\_flowers})-(\text{total\_price\_3\_flowers}+\text{total\_price\_2\_flowers})

This line calculates the amount of money saved by buying at the better price, it's correct.

Overall, the code correctly calculates the amount of money saved by buying 18 flowers at the better price.
& \\

\textbf{Here's the better code:} \\
\text{num\_flowers} = 18 \\
\text{price\_3\_flowers} = 2.5 \\
\text{price\_2\_flowers} = 1 \\

\text{total\_price\_3\_flowers} = (\text{num\_flowers} // 3)*\text{price\_3\_flowers} \\
\text{remaining\_flowers} = \text{num\_flowers} \% 3 \\
\text{total\_price\_2\_flowers} = (\text{remaining\_flowers}//2)* \text{price\_2\_flowers} \\

\text{money\_saved} = (\text{num\_flowers} // 3) * \text{price\_3\_flowers} + (\text{num\_flowers} \% 3) * \text{price\_2\_flowers} - \text{num\_flowers} * \text{price\_2\_flowers} \\

\text{answer} = \text{abs}(\text{money\_saved})\\
\textbf{Extract the Answer:} 3.0 
(\textbf{\textcolor{red}{\ding{55}}}) Code 1&\\

&\\
\textbf{Instruct the Iteration (Refresh):}&\\
The error lies in both code segments not correctly implementing the meta-thought's guidance to compare the total costs of each purchasing option, ultimately resulting in incorrect calculations of Vincent's savings.
Decision: Refreshing the response.&\\
\hline
\textbf{Refresh}&\multirow{8}{=}{\centering Iteration 2\\6.0 (\textcolor{green}{\ding{51}})}:\\
\text{num\_flowers} = 18 \\
\text{price\_3\_flowers} = 2.5 \\
\text{price\_2\_flowers} = 1 \\

\text{cost\_using\_3\_flower\_packages} = (\text{num\_flowers} / 3) * \text{price\_3\_flowers} + (\text{num\_flowers} $\mod$ 3) * \text{price\_2\_flowers} \\
\text{cost\_using\_2\_flower\_packages} = (\text{num\_flowers} / 2) * \text{price\_2\_flowers} + (\text{num\_flowers} $\mod$ 2) * (\text{price\_2\_flowers} / 2) \\

\text{money\_saved} = \text{cost\_using\_2\_flower\_packages} - \text{cost\_using\_3\_flower\_packages} \\

\text{answer} = \text{money\_saved}\\
\textbf{Extract the Answer:} 6.0 
(\textcolor{green}{\ding{51}}) Code 2&\\
&\\
\textbf{Instruct the Iteration (Select)}:& \\
COT 2 aligns better with the meta-thought of accurately calculating total costs for each buying strategy. Code 2 indicates a greater benefit from selecting the optimal option, ensuring Vincent maximizes his savings on flowers. Better Code after comparison: Code 2\\
\hline

\caption{An illustrative example from GSM8K demonstrating how IoRT generates a refresh instruction when the results before and after reflection are consistent.}
\label{table:case_study8}
\end{longtable}
\twocolumn

\end{document}